# Analyzing and Encoding the Al-Mawrid Arabic-English Dictionary with the ISO Language Markup Framework and TEI Lex-0

**Diaa M. Fayed**[1] **Laurent Romary**[2]

[1]Faculty of Information Technology and Computer Science,
Sinai University, Arish, Egypt,
https://orcid.org/0000-0002-6330-6328
diaa.fayed@su.edu.eg, diaa.fayed@outlook.com

[2]Inria, Director for Scientific Information and Culture,
ISO committee TC 37 (Language and terminology), president (2016-),
DARIAH EU infrastructure, president of the Board of Directors (2014-2018),
https://orcid.org/0000-0002-0756-0508
laurent.romary@inria.fr

**Abstract**. This paper presents a robust methodology for the systematic digitization and encoding of the *Al-Mawrid* Arabic–English dictionary, transforming it from a legacy print resource into a standardized computational lexicon. Addressing a significant gap in Arabic lexical infrastructure, the study adopts a dual-standard framing that aligns the ISO Lexical Markup Framework (LMF) with the Text Encoding Initiative TEI Lex-0 guidelines. By applying an editorial view to the dictionary's macro- and microstructure, the research resolves the structural ambiguities and punctuation inconsistencies typical of 20th-century bilingual dictionaries.

The methodology is grounded in an empirical analysis of the dictionary's lexical knowledge density. Drawing on a representative sample (the letter Ayn, comprising 4.6% of the total volume), the study provides scientific weight to the encoding process, demonstrating a structural parsing accuracy of 91%. Quantitative evaluation of the information extraction rules reveals high performance, with 85% precision and 98% recall for synonyms, and 88% precision for other morpho-semantic features.

Beyond technical description, the paper provides a critical comparison with existing Arabic lexical resources and discusses the limitations of TEI Lex-0 when modelling specific Arabic phenomena, such as implicit "open set" semantic relations and scattered morphological cues. Furthermore, the study explores the potential for Linguistic Linked Open Data (LLOD) integration by establishing a scalable prefix-based referencing system that facilitates the resource's inclusion in the semantic web. The result is an interoperable, machine-tractable resource that provides a reproducible workflow for the retro-digitization of complex legacy bilingual lexicons within the Arabic NLP and Digital Humanities communities.

## 1. Introduction

Despite a rich and long-standing lexicographical tradition, the Arabic language suffers from a significant scarcity of lexical resources encoded according to international interoperability standards. While widely used reference works like the Al-Mawrid Arabic-English dictionary provide vast linguistic data, they remain largely "locked" in print-oriented or inconsistent digital formats that are inaccessible to modern computational tools. This paper addresses this research gap by proposing a standardized methodology to transform the *Al-Mawrid* into a machine-readable resource, facilitating its integration into Natural Language Processing (NLP) and Digital Humanities applications

To achieve this, we employ a dual-standard framework. The ISO Language Markup Framework *series* (LMF, ISO 24613) provides the essential structural model for representing complex semasiological lexical data. This is complemented by the Text Encoding Initiative (TEI) guidelines and, specifically, TEI Lex-0, a community-driven baseline that streamlines the encoding of retro-digitized dictionaries. By aligning these two frameworks, this study demonstrates a methodology that preserves the dictionary's macro- and microstructural integrity while enhancing its technical interoperability.

In the transition from print to digital, we adopt the editorial view Costa *et al.*(2022), Ide and Véronis (1995). Unlike the typographic view, which prioritizes visual layout, the editorial perspective treats the dictionary as a meaningful linear stream of tokens, allowing us to focus on the systematic representation of typographic, syntactic (dictionary structure) and semantic (identifying meaningful components in an entry) information. This approach is particularly effective for resolving the structural ambiguities and implicit linguistic cues—such as those found in multiword expressions (MWEs) and cross-references—that are prevalent in *Al-Mawrid*.

The primary contributions of this work are threefold:

1. The production of a validated TEI Lex-0/LMF encoding of the *Al-Mawrid* suitable for computational use.
2. The resolution of structural inconsistencies typical of legacy print dictionaries through standardized markup.
3. The proposal of a reusable workflow for the retro-digitization of other Arabic legacy dictionaries, ensuring they meet FAIR (Findable, Accessible, Interoperable, and Reusable) data Principles Wilkinson *et alI* (2016).

The remainder of the paper is structured as follows: Section 2 surveys relevant literature on Arabic lexical modelling; Section 3 provides a structural analysis of the *Al-Mawrid*; Section 4 details the core encoding methodology with illustrative examples; Section 5 discusses results and evaluation; and Section 6 concludes with future directions for Linked Open Data (LOD) integration.

## 2. Related Works

Research on encoding Arabic lexical resources has evolved significantly over the past two decades, transitioning from basic digitisation to complex ontological modelling. Despite this progress, a critical gap remains: large Arabic–English legacy bilingual dictionaries are fundamentally under-represented in standardised digital form.

The *Al-Mawrid* dictionary is a critical linguistic resource because it serves as one of the most widely used bilingual references in the Arabic-speaking world. For non-experts, its importance lies in its status as a prominent landmark of Arabic lexicography, acting as a vast repository of lexical knowledge that bridges the gap between Arabic and English for diverse applications.

Notably, the *Al-Mawrid* dictionary has never been digitised or modelled in prior research. To identify the research gap filled by this study, in this section, we tackle the analysis of the state of the art of Arabic digital lexicography under four complementary angles: (1) LMF-based modelling of Arabic legacy resources; (2) the application of standards to complex Arabic phenomena; (3) the evaluation and enrichment of standardised lexica; and (4) the emerging alignment between LMF and TEI Lex-0.

### 2.1 LMF-based Modelling and Retro-digitisation of Arabic Legacy Resources

The foundation for standardised Arabic lexica was established by Salmon-Alt et al. (2005), who introduced a normalised representation to account for morphological complexity. Subsequent efforts focused on the transformation of printed dictionaries into machine-readable formats using the ISO Lexical Markup Framework (LMF). This includes Loukil et al. (2008, 2010), who developed syntactic models for Arabic verbs, and Ammar et al. (2008), who proposed a generalised data model for MRDs. A comprehensive normalisation pipeline was later introduced by A. Khemakhem *et al*. (2009–2015), validating the conversion of the *Al-Ghani* dictionary for use in modern query systems. Furthermore, Graff and Maamouri (2012) demonstrated the utility of LMF in modernising bilingual dictionaries for colloquial Arabic dialects to ensure cross-resource consistency.

## 2.2 Standardised Modelling of Complex Arabic Phenomena

To handle the inherent complexity of the Arabic language, the LMF framework has been adapted to model specific linguistic phenomena. This includes the development of WordNets for Tunisian Arabic (Karmani et al., 2014; Moussa & Alimi, 2015) and the compilation of lexica for Arabic particles (Namly et al., 2015). This line of research highlights how the framework can be extended to handle non-standard varieties and specific part-of-speech categories. Building on these foundations, recent work has explored more nuanced Arabic lexical representations in TEI (cf. Nahli & Del Grosso, 2021) and adaptations for handling Arabic complexity (cf. Nahli et al., 2016).

## 2.3 Interoperability, Evaluation, and Enrichment of Lexical Resources

A significant area of study involves how LMF encodings are operationalised and evaluated within broader systems. Ben Abderrahmen et al. (2009) developed graphical tools to extract complex information from LMF databases, while Wali et al. (2014) proposed systems to validate these dictionaries against Data Category Registries (DCRs). Interoperability is further enhanced through the generation of ontologies from standardised dictionaries to support Information Retrieval and Word Sense Disambiguation (Amar et al., 2010; Soudani et al., 2014). These studies emphasise that standardised encoding is not an end in itself but a prerequisite for semantic enrichment and system integration.

## 2.4 TEI Lex-0 Retro-digitisation and LMF–TEI Alignment

The current paper's primary value lies in its dual framing, aligning ISO-LMF structural models with TEI Lex-0 serialisation. Prior work by Haddar et al. (2012) demonstrated the feasibility of converting syntactic lexica into LMF and subsequently serialising them into TEI. Similarly, Elleuch et al. (2013, 2015) utilised a rule-based approach to enrich LMF-normalised dictionaries with syntactic grammars before TEI serialisation. Despite these advancements, there remains a lack of studies providing a TEI Lex-0 encoding explicitly aligned with LMF for large-scale legacy bilingual dictionaries. This work fills that specific gap by applying this dual perspective to *Al-Mawrid*, ensuring both structural rigour and community-standard interoperability.

## 2.5 Comparative Analysis of Standardised Arabic Lexical Modelling

While the preceding sections demonstrate a robust history of Arabic lexicon modelling, a significant research gap remains regarding the application of modern, community-driven standards to large-scale legacy bilingual resources. Much of the existing literature focuses on specialized lexica—such as verb-specific syntactic models, particle lexicons, or religious-domain ontologies—leaving general-purpose bilingual dictionaries like the Al-Mawrid largely under-represented in standardized digital forms.

Drawing from the related works, Table 1 provides a systematic comparison between the current study and key existing Arabic lexical resources encoded under international standards.

Table 1: Comparison of Standardised Arabic Lexical Modelling Studies.

| Research Work | Resource Type | Standard Employed | Primary Focus / Methodology |
|---|---|---|---|
| Current Study | *Al-Mawrid* (Arabic–English) | LMF & TEI Lex-0 | Dual-standard alignment; resolution of structural ambiguities in legacy bilingual resources. |
| A. Khemakhem et al. (2015) | *Al-Ghani* (Monolingual) | LMF (ISO 24613) | Comprehensive normalisation pipeline; validation against DTDs and DCRs. |
| Graff & Maamouri (2012) | Colloquial Dialects (Bilingual) | LMF | Unified format for interoperability across various colloquial Arabic dialects. |

| Haddar et al. (2012) | HPSG Syntactic Lexica | LMF & TEI | Rule-based conversion and serialisation of syntactic lexica. |
|---|---|---|---|
| Elleuch et al. (2015) | Monolingual Lexica | LMF | Syntactic enrichment through corpus-based grammars using the NOOJ platform. |
| Soudani et al. (2014) | Hadith Dictionaries | LMF | Normalisation of unstructured religious texts to build semantic ontologies. |

Unlike previous works that focused primarily on the LMF abstract model for monolingual dictionaries (e.g., A. Khemakhem et al., 2015), this study addresses unique challenges in cross-linguistic alignment and Translation Equivalence Groups (TEGs) for a general-purpose bilingual dictionary. By adopting an editorial view, the study prioritises the faithful conversion of the dictionary's linear stream of tokens into a structured format that preserves original lexicographic depth while enhancing machine-readability.

## 3. Material and Structural Analysis

Before initiating the encoding workflow, a rigorous analysis of the source material's typographic and structural properties is required to bridge the gap between print layout and digital schema. Al-Mawrid employs a complex system of typographic cues—such as bolding, indentation, colons, comma, and semicolons—to signal semantic and syntactic relationships. This section provides the necessary background on the dictionary's macrostructure (headword arrangement) and microstructure (the internal logic of senses and Translation Equivalence Groups), establishing a baseline for the heuristic interpretations used in the subsequent methodology.

### 3.1 Macrostructural Organisation

The macrostructure of *Al-Mawrid* is governed by a traditional alphabetical arrangement following the conventions of the Arabic script. Quantitative analysis of (Baalbaki, 2004; Fayed et al., 2014) reveals that the dictionary contains a substantial volume of lexical data, comprising approximately 35,000 distinct headwords and roughly 60,000 subentries.

- Primary Reference Units: Boldface headwords function as the core nodes of lexical information.
- Semantic Grouping: Individual senses tied to a specific headword are presented as nested subentries.
- Visual Hierarchies: Typographic cues, such as left-indentation on new lines from the second sense onward, visually differentiate meanings associated with a single lemma.

### 3.2 Microstructural Organisation

The microstructure typically consists of a mandatory Arabic portion and an optional English translation section, each following a specific internal logic.

- Headword Types: While most entries feature single-word headwords, the dictionary frequently includes "head-phrases" comprising more than one word.
- Arabic Section Fields: This portion is subdivided into three potential fields: a header (required), an explanation, and a cross-reference (both optional).
- Translation Equivalents: The English section is organised into Translation Equivalence Groups (TEGs). Commas separate synonymous equivalents, while semicolons define boundaries between distinct shades of convergent meanings or subsenses.

### 3.3 Quantitative Distribution and Parsing Regularity

To provide the necessary scientific weight, a quantitative study (Fayed *et al.*, 2014) was conducted on a representative sample of the Ayn (ع) letter, which constitutes 4.6% of the volume. Analysis of the microstructure's punctuation regularity reveals a structural parsing accuracy of 91%. By mapping these scattered cues—approximately 13% of which are cross-references—into the dual-standard framework, the methodology resolves the "hidden" hierarchies of the printed page into a fully navigable digital network. The high degree of punctuation regularity (91%) observed in this 4.6% pilot sample indicates that the heuristic transformation rules are highly reproducible and scalable to the remaining volume of the dictionary with minimal adjustment.

### 3.4 Linguistic Markers and Punctuation Regularity

The transition from a print-oriented layout to a machine-actionable structure necessitates a rigorous formalisation of the dictionary's visual delimiters. Under the editorial view, punctuation marks and labels are treated as a linear stream of functional tokens that define the boundaries between lexical, semantic, and syntactic fields. To establish the foundational logic for the subsequent automated parsing process, it is necessary to identify the markers that signify linguistic metadata. Table 2 provides a summary of the primary Arabic labels and abbreviations used to classify the dictionary's lexical content

Table 2: Summary of Arabic Labels and Abbreviations.

| **Label / Abbreviation** | **Meaning** |
|---|---|
| [طب]، [كيمياء] | Medicine, Chemistry |
| (نبات)، (حيوان) | Plant, Animal |
| ج / مفردها | Plural / Singular |
| راجع | Refer / Cross-reference |
| ضد / لا / غير | Antonym markers |
| مصدر | Infinitive / *Maṣdar* |

Punctuation marks in *Al-Mawrid* fulfil critical structural and interpretive roles, defining the boundaries between headers, definitions, and translations. As illustrated in the microstructural snapshot of the letter Ayn (Figure 1), a colon (:) typically separates the header from the explanation, while Arabic commas (،) or the conjunction "أو" (or) are used to separate synonymous words or phrases. Within the English section, semicolons distinguish between different Translation Equivalence Groups (TEGs), separating shades of convergent meaning that are not strictly synonymous. Table 3 systematises these structural roles to guide the heuristic transformation of the text.

.

Table 3: Punctuation Markers and Structural Roles.

| **Symbol** | **Location** | **Function** |
|---|---|---|
| Colon (:) | Arabic | Introduces a definition or explanation. |
| Comma (،) | Arabic | Separates synonyms or orthographic variants. |
| Dash (–) | Arabic | Precedes a cross-reference label. |
| Semicolon (;) | English | Separates English subsenses/translation groups. |
| Comma (,) | English | Separates synonymous translation equivalents. |
| Parentheses ( ) | Both | Enclose contextual glosses or grammatical info. |
| Brackets [ ] | Arabic | Specifically enclose semantic domain labels. |

The presence of high punctuation regularity—which enabled a structural parsing accuracy of 91% in independent error source analysis—validates the adoption of this editorial perspective for digital encoding. By mapping these scattered cues into the LMF/TEI Lex-0 framework, the methodology resolves the "hidden" hierarchies of the printed page into a fully navigable digital network

3.5 Quantitative Distribution of Lexical Knowledge

The scientific weight of the proposed methodology depends on providing evidence-based indicators of the resource's coverage and extraction accuracy. To provide context for the pilot scale of our encoding, we incorporate baseline statistics from Fayed *et al*. (2014) regarding the overall variety of linguistic data categories available in the source material. Utilizing a representative sample of the Ayn (ع) letter chapter—representing 4.6% of the dictionary's total volume—Table 4 quantifies the distribution of significant lexical relations and evaluates the extraction performance of the heuristic rules. In this context, precision signifies the accuracy of the extracted relations (the proportion of correct results among retrieved items), while recall measures the system's coverage (its ability to identify all relevant relations present in the source material).

Table 4: Distribution and Extraction Performance of Lexical Relations.

| Relation Type | Significance | Signal Type | Precision | Recall |
|---|---|---|---|---|
| Synonyms | **Highest** | Implicit (Commas) | 85% | **98%** |
| Domain Labels | High | Explicit (Brackets) | 88% | 73% |
| Derivations | High | Explicit (*Maṣdar*) | 88% | 73% |
| Hyponyms | Moderate | Implicit (Genus) | 88% | 33%* |
| Antonyms | Low | Explicit (*Ḍidd*) | 88% | 73% |

**Note*: The low recall for hyponyms is attributed to the fact that 77% of these relations in the source material are implicit "open set" structures that follow a genus-differentia pattern rather than utilizing fixed keywords.

The evaluation reveals that while synonymous relations are highly regular and effectively retrieved, hierarchical relations such as hyponymy are frequently embedded within implicit natural-language phrasing. This underscores a major limitation of applying standard schemas to legacy Arabic dictionaries: the near-total lack of formal tagging requires the encoder to rely on "expert induction" to infer and then formalise these relations.

To establish a baseline for structural parsing and regularity, we draw on quantitative findings by Fayed *et al*. (2014), which recorded a structural parsing accuracy of 91% when accounting for independent error sources. This high degree of regularity validates the decision to adopt an editorial view for digital encoding, as the primary sources of error stem from in-dictionary inconsistencies (such as missing colons or mismatched headwords) rather than failures of the parser logic. Table 5 provides a granular breakdown of these error sources and their relative frequency within a representative sample.

Table 5: Independent Error Analysis of Structuring Steps.

| Error Source | Frequency | Example Failure |
|---|---|---|
| In-Dictionary | **Highest** | Missing colons or inconsistent domain brackets. |
| Data Entry | Moderate | Entries dropped or shifted during digitisation. |
| Parser | **Lowest** | Faulted characters or unexpected whitespace. |

By mapping these scattered linguistic cues - approximately 13% of which are cross-references - into the dual-standard framework, the methodology resolves the "hidden" hierarchies of the printed page into a fully navigable digital network. This transition ensures that the resource is prepared for immediate integration into automated Arabic NLP pipelines and Linguistic Linked Open Data (LLOD) frameworks.

To demonstrate the practical application of these structural principles, Figure 1 presents the full entry for the headword ʿayn (عَيْن). This entry provides a representative sample of the dictionary's microstructural complexity, containing a mandatory Arabic section and various optional fields such as explanations, cross-references, and Translation Equivalence Groups (TEGs).

The fundamental structural components of the Al-Mawrid subentries have been previously identified through linear analysis (Baalbaki, 2004; Fayed et al., 2014), which defines the hierarchy of mandatory headers and optional cross-references.

By adopting an editorial view for this analysis, we prioritise the systematic representation of typographic and semantic cues – such as indentation and colons - to resolve the structural ambiguities inherent in legacy print formats. This approach ensures that the "hidden" hierarchies of the printed page are transformed into an explicit, machine-actionable dictionary.

| eye | **عَيْن**: عُضْوُ البَصَر | |
|---|---|---|
| evil eye | عَيْن، عَيْنٌ لامَّة ، عَيْنٌ مُصِيبَةٌ بِسُوء | |
| spring, source, fountainhead, headspring, spring, wellhead, wellspring, springhead, fountain, well | عَيْن: يَنْبُوع | |
| spy | عَيْن: جاسُوس | |
| scout, reconnoiterer, explorer | عَيْن: مُسْتَكْشِف، مُسْتَطْلِع | |
| overseer, inspector, supervisor, superintendent | عَيْن: ناظِر، مُرَاقِب | |
| lookout, guard, watch(man) | عَيْن: حارِس | |
| eye, threading hole | عَيْن: ثُقْبُ الإبْرَة | |
| specie, coin(s) | عَيْن: عُمْلَةٌ مَعْدِنيَّة | |
| cash, ready money | عَيْن: نَقْدٌ حاضِر | |
| property, assets, personality, personal property, chattel(s), goods | عَيْن: مال، أَمْوال، مِلْك | |
| usury, interest | عَيْن: رِبَا، فائِدَة | |
| choice, pick, prime, flower, cream, top, elite | عَيْن: خِبْرَة، صَفْوَة | |
| notable, dignitary, person of distinction, prominent person, leading personality | عَيْنٌ: وَجِيهٌ | |
| lord, master, chief, head | عَيْن: سَيِّد | |
| substance, essence | عَيْن: جَوْهَر | |
| the same, the very, the selfsame, the identical | عَيْن (كذَا)، عَيْنُهُ، بِعَيْنِهِ: نَفْسُهُ، ذاتُهُ | |
| opal | عَيْنُ الشَّمْس: أُوبَال (حَجَرٌ كَرِيم) | |
| second radical of a verb | عَيْنُ الفِعْل [لغة] | |
| cat's-eye | عَيْنُ الهِرّ: حَجَرٌ نِصْفُ كَرِيم | |
| in kind, in specie | عَيْنًا | |
| an eye for an eye, tit for tat | العَيْنُ بِالعَيْن | |
| with the naked (or unaided) eye | بِالعَيْنِ المُجَرَّدَة | |
| none other than, the very, the very same, the same, the selfsame, the identical; personally, in person | بِعَيْنِه، هُوَ عَيْنُهُ | |
| | أَخَذَ بِعَيْنِ الاعْتِبَار – راجع أَخَذَ | |
| concrete noun | اسْمُ عَيْنٍ [لغة] | |
| with one's own eyes | بِأُمِّ عَيْنِهِ ، بِعَيْنِهِ | |
| to see with one's own eyes | رَأَى رأْيَ العَيْن | |
| | طَرْفَةُ عَيْن – راجع طَرْفَة | |
| | على الرَّأْسِ والعَيْن – راجع رَأْس | |
| individual duty | فَرْضُ عَيْن | |
| | مَجْلِسُ أَعْيان – راجع مَجْلِس | |

Figure 1: The entry "عَيْن" in the Al-Mawrid.

## 4. Encoding Methodology: The Dual LMF/TEI Lex-0 Framework

The transformation of *Al-Mawrid* from a print-based reference into a machine-readable resource requires a methodology capable of resolving high levels of structural ambiguity and implicit linguistic data. To ensure both theoretical rigor and technical interoperability, this study adopts a dual-standard approach: we utilize the ISO Language Markup Framework (LMF, ISO 24613) for its robust conceptual modeling of lexical structures and TEI Lex-0 for its streamlined, community-standardized XML serialization. This alignment is critical because, while TEI provides the vocabulary for representational encoding, the LMF layer ensures that the underlying lexical model remains compliant with international standards for data exchange and NLP integration.

The encoding methodology for *Al-Mawrid* is specifically designed to resolve the structural ambiguities inherent in its legacy print-based lexicography. Unlike modern digital-born resources, Al-Mawrid relies heavily on typographic cues—such as indentation and punctuation—to signal semantic boundaries, while largely lacking explicit part-of-speech (POS) tags. Consequently, our workflow is not a mere tag-assignment exercise but a heuristic transformation governed by the editorial view

Our encoding strategy is governed by the editorial view, which treats the dictionary as a linear stream of tokens. This perspective allows us to systematically map the dictionary's unique features—such as indented subentries and implicit part-of-speech (POS) categories—into explicit, queryable XML structures. The following subsections detail the encoding of key components, including entries, multiword expressions (MWEs), and cross-referencing schemes. Moving beyond a purely technical description, this methodology section focuses on the research motivations and the "why" behind each modelling decision—specifically how the constraints of TEI Lex-0 and the requirements of LMF guided the resolution of specific Arabic lexicographic challenges.

To support the "dual-standard" claim, the conceptual mapping between the two frameworks is made explicit in the table below, serving as the foundational logic for the rules and illustrations that follow. Table 6 provides a foundational mapping for the subsequent methodology, clarifying how specific LMF conceptual classes concretely shape our modelling choices within the TEI Lex-0 framework and highlighting the research motivation behind each mapping decision.

Table 6: Mapping of Al-Mawrid Methodology to LMF and TEI Lex-0.

| **LMF Class (Conceptual Layer)** | **TEI Lex-0 Element** | **Methodological Function in Al-Mawrid** |
|---|---|---|
| LexicalEntry | `<entry>` | Represents the primary reference unit or head-phrase,,. |
| LexicalEntry (Nested) | `<entry type="homonymicEntry">` | Disambiguates headwords functioning across multiple parts of speech. |
| RelatedEntry | `<entry type="relatedEntry">` | Models MWEs and subentries that carry their own semantic data. |
| Lemma | `<form type="lemma">` | Encapsulates the orthographic representation of the Arabic headword. |
| Form (Variant) | `<form type="variant">` | Preserves orthographic lemma variations found in the entry header. |
| Form (Inflection) | `<form type="inflected">` | Represents plurals or specific inflections signaled in parentheses. |
| Form (Derivative) | `<form type="derived">` | Models derived forms like *maṣdar* (infinitive) markers. |
| Sense | `<sense>` | Models individual meanings or semicolon-delimited subsenses. |
| Equivalent | `<cit type="translationEquivalent">` | Maps Arabic source words to English target-language equivalents. |
| Equivalent (Free text) | `<cit type="translation">` | Used when the translation is a descriptive phrase rather than a direct |

| | | equivalent. |
|---|---|---|
| MorphosyntacticProperties | <gramGrp> | Formalises implicit or inferred POS and grammatical cues. |
| SubjectField | <usg type="domain"> | Converts square-bracketed domain labels into queryable metadata. |
| Context | <gloss> | Captures parenthetical clarity markers and unstructured semantic cues. |
| SenseRelation | <xr> | Functions as a structural wrapper for the entire cross-referencing block. |
| (Implicit Link) | <ref> | Provides the machine-actionable target URI for cross-references. |
| Statement (Label) | <lbl> | Formalises abbreviated cues for gender, number, or cross-referencing keywords. |
| Example | <cit type="example"> | Encapsulates illustrative sentences demonstrating lemma usage. |
| (Text Content) | <quote> | Holds the actual textual content of illustrative examples or free translations. |
| (Structural Marker) | <pc> | Formalises punctuation marks used as functional delimiters between fields. |

To illustrate these principles, the subsequent sections employ a minimal, representative set of "miniature proofs" that focus on the methodological logic and research motivations behind each modelling choice. Each example serves as a functional demonstration of how print inconsistencies are transformed into structured, machine-readable data without compromising the integrity of the original resource. While this section prioritises the "why" behind specific encoding decisions to ensure clarity, the full breadth of the dictionary's patterns—including rich and repetitive examples—is provided in Appendices A through K, organised by subsection topic.

## 4.1 Entry Representation and Lemma Modelling

The encoding of *Al-Mawrid* centers on the <entry> element, which serves as the primary structural unit for aggregating all lexical, semantic, and syntactic information associated with a specific lemma. In the original print edition, an entry is defined by a boldface header and often spans multiple lines, using visual indentation to distinguish subsequent senses. Our methodology translates this visual hierarchy into a machine-readable format by adopting the editorial view, which represents these indented lines as distinct <sense> blocks; specifically, this approach utilizes nested <sense> elements to explicitly model the subsenses signaled by semicolons in the source material.

To satisfy the interoperability requirements of the ISO-LMF standard while adhering to the TEI Lex-0 baseline, every <entry> is assigned a unique xml:id and a mandatory xml:lang attribute. These are not merely technical prerequisites; they are methodological solutions to the structural ambiguities of legacy dictionaries, ensuring that every headword and its associated senses are computationally addressable and unique within a global NLP pipeline.

The transformation of the dictionary's primary reference unit is best illustrated by the following example in Figure 1 where a single-word lemma and its translation are aligned within the <entry> framework. This mapping satisfies the structural requirements of TEI Lex-0 while ensuring that the headword remains computationally addressable and unique within a global NLP pipeline.

| Full/Partial Entry | أَبْضِيّ popliteal |
|---|---|
| LMF/TEI Lex-0 Encoding | <entry xml:id="me-000001" type="mainEntry" xml:lang="ar"><br><form type="lemma"><orth>أَبْضِيّ</orth></form><br><sense n="se-01"><br><cit type="translationEquivalent" xml:lang="en"><br><form><orth>**popliteal**</orth></form></cit></sense></entry> |

Figure 2: Simplest entry: single-word lemma and single-word translation.

For a comprehensive collection of entry types, including multi-lemma headers and prepositional phrases, refer to Appendix A: Entry Representation and Lemma Modelling.

### 4.1.1 Grammatical Mapping and Header Typolog

A fundamental challenge in *Al-Mawrid* is the near-total lack of formal part-of-speech (POS) tags, which occur in less than 1% of entries. To resolve this, our methodology adopts a heuristic approach where the structural category of the header dictates the application of explicit linguistic metadata via the <gramGrp> and <gram> elements. This approach is applied across three primary header types to ensure the resource is computationally actionable:

- Single-Word Lemmas: For simple entries, the orthographic headword is encapsulated within <form type="lemma">. While the POS is often implicit, this structural isolation allows for the manual or automated attachment of inferred tags—such as "Noun" or "Verb"—inside a <gramGrp> block, facilitating future automated indexing.

- Prepositional Collocates: When a lemma is paired with a specific preposition (e.g., *ṣarraḥa bi-*), we use a <gram type="collocate"> element within a <gramGrp> container. This decision is motivated by the need to formalise syntactic dependencies that are only implicitly suggested in print, thereby enhancing the resource's utility for modern syntactic parsers.

- Orthographic Variants: Headers containing spelling variations (e.g., *ʿajufa* and *ʿajifa*) are modelled as nested <form type="variant"> elements. By nesting these within a parent lemma, we ensure they share the same inferred grammatical properties (such as plurality or derivation type) formalised in the parent's <gramGrp>, preserving microstructural integrity while enabling precise lexical search.

### 4.1.2 Modelling Multiword Expression (MWE) Headers

The most significant microstructural complexity occurs when a header functions as a Multiword Expression (MWE). For macrostructural MWEs—those that serve as primary headwords—the encoding mirrors the single-word lemma structure but is enriched with an LMF-compliant <gram type="mwe"> element.

The inclusion of the @value attribute (e.g., value="N-N") is a strategic choice to make the LMF layer explicit. By specifying the parts of speech of the MWE’s components, we provide the high-level linguistic specifications required by ISO 24613 (*part* 2), ensuring that the digital *Al-Mawrid* transcends its origins as a print reference to become a robust, standardized dataset for Arabic NLP applications.

## 4.2 Labels, Punctuation, and Semantic Marking

The transformation of *Al-Mawrid* into a machine-readable resource requires a systematic approach to the dictionary’s vast array of labels and punctuation, which serve as the primary visual cues for structural and interpretive boundaries. Our methodology prioritises structural disambiguation, transforming these print-oriented indicators into explicit metadata that supports automated parsing and domain-specific filtering.

#### 4.2.1 Domain and Linguistic Labels

To facilitate diatechnical marking—a critical requirement for domain-specific NLP applications—we map *Al-Mawrid's* field indicators (e.g., [طب] for medicine or [فلك] for astronomy) to the TEI <usg type="domain"> element. This decision is motivated by the need to ensure that semantic fields are not merely preserved as text but are functionally queryable within a digital lexicon.

Furthermore, the dictionary employs alphabetical indicators to convey grammatical categories (such as ج for plural or مفردها for singular). While these are visually distinct in print, our encoding formalises them using the <lbl> (label) element. This creates a structured link between the abbreviated label and its underlying linguistic category, directly supporting the ISO-LMF goal of providing high-level specifications for lexical content. The following example in Figure 3 shows the conversion of a visual marker into searchable metadata.

| **Label** | **Translation** | **Encoding** |
|---|---|---|
| ]رياضة[ | Sports | `<pc>[</pc><usg type="domain">رياضة</usg><pc>]</pc>` |

Figure 3: Domain label encoding.

A complete registry of Arabic labels, abbreviations, and their functional structural roles is provided in Appendix B: Labels, Punctuation, and Semantic Marking

#### 4.2.2 Punctuation and Interpretive Boundaries

Punctuation in *Al-Mawrid* is more than a typographic feature; it defines the functional relationship between linguistic objects. We employ the <pc> (punctuation) element to formalise these delimiters, following a strict logic based on their interpretive roles:

- Colons (:) are encoded to signal the transition from a header to a definition or explanation, essentially marking the boundary between the lemma and its semantic interpretation.

- Commas (,) are used to distinguish between orthographic variants or synonymous phrases, enabling NLP tools to identify discrete lexical items within a single entry.

- Parentheses (...) present a specific modelling challenge, as they contain both structured (syntax/semantic knowledge) and unstructured information. Following the editorial view, we encode structured parenthetical data into standard TEI elements while preserving unstructured "clarity" cues within <gloss> elements.

By formalising these cues, we resolve the implicit relationships found in the print edition, ensuring that the digital resource adheres to FAIR data principles by making the dictionary's internal logic accessible to machine-processing tools.

### 4.3 Lexical Structure and Reference Modelling

Standardising the architecture of *Al-Mawrid* requires a formal mapping of its visual macrostructure and microstructure into a machine-actionable hierarchy. Our methodology translates the print dictionary's reliance on typography and space into a recursive XML structure, ensuring that the implicit relationships between headwords and senses are explicitly defined for computational use. To establish connectivity, this fragment in Figure 4 shows how structural roles are encoded within unique IDs.

| Structural Component | LMF/TEI Lex-0 XML Fragment |
|---|---|
| Main Entry | `<entry type="mainEntry" xml:id="me-000001">` |
| Sense | `<sense n="se-01">` |

Figure 4: Macrostructure, Microstructure and Referencing system.

Detailed diagrams of the recursive XML hierarchy and the zero-padded ID registry are available in Appendix C: Lexical Structure and Reference Modelling.

4.3.1 Macrostructure: From Print Layout to XML Hierarchy

In its original form, *Al-Mawrid* follows the alphabetical conventions of the Arabic script, utilizing visual indentation to group related senses under a single boldface headword. To preserve this semantic grouping while enabling high-granularity data retrieval, we represent each headword as a top-level <entry> element.

The motivation for our nesting strategy lies in the resolution of semicolon-delimited translations. In the print edition, semicolons are used to separate convergent meanings that are not strictly synonymous. Following the TEI Lex-0 baseline, we parse these semicolon-delimited groups into distinct <sense> or nested <sense> elements. This methodological choice transforms a flat list of equivalents into a layered semantic model, allowing Natural Language Processing (NLP) tools to distinguish between subtle shades of meaning that were previously "locked" within a single typographic line.

4.3.2 Microstructure: Disambiguating Homonyms and Multiword Expressions

The microstructure of *Al-Mawrid* is characterized by a high frequency of "head-phrases" and lemmas serving multiple grammatical roles. Our encoding methodology addresses these complexities through a dual-entry classification system:

• Homonymic Disambiguation: When a headword functions under multiple grammatical categories (e.g., as both a noun and a verb), we encode each role as a nested <entry type="homonymicEntry">. The reason for this nesting, rather than using flat sibling entries, is to maintain the integrity of the lemma while providing the fine-grained linguistic analysis required for automated disambiguation tasks.

• MWE Integration: To distinguish between core lexical senses and fixed sequences, we categorize subentries into direct senses (<sense>) and related entries (<entry type="relatedEntry"> for Multiword Expressions). This layered structure is a strategic intervention designed to support clear semantic organization; it ensures that idiomatic expressions are not lost within the definition text but are instead treated as unique lexical objects with their own specific IDs.

By formalizing these structural boundaries, the methodology ensures that the digital resource does not merely mirror the print layout but actively enhances its utility for Digital Humanities and NLP applications by making the dictionary's internal logic fully explicit.

4.3.3 Systematic Referencing and Identification

A fundamental challenge in transitioning *Al-Mawrid* from a print-based reference to a machine-readable resource is the lack of formal regularity in its original layout. To resolve these structural ambiguities and ensure the resource is suitable for Natural Language Processing (NLP) and Digital Humanities (DH) applications, we have implemented a scalable referencing system based on unique identifiers (@xml:id) and sequential values (@n).

The motivation for this systematic identification is twofold: first, to enable precise internal cross-referencing within a complex macrostructure; and second, to facilitate external computational access to specific lexical objects. Rather than adopting a flat numbering scheme, we employ a prefix-based taxonomy that explicitly encodes the structural role of each component within the ID itself. This design choice ensures that the digital architecture remains both human-interpretable and machine-parsable:

• Entry Classification: Primary headwords are distinguished by the prefix "me" (main entry), while hierarchical complexities like homonyms and subentries are clarified via "hm" (homonymic entry) and "re" (related entry).

• Semantic Granularity: To support fine-grained linguistic analysis, distinct identifiers are assigned to semantic layers, with "se" designating main senses and "su" identifying the subsenses derived from semicolon-delimited groups in the print text.

By adopting a standardised format—utilising six-digit numeric codes for entries and zero-padded two-digit values for senses—the methodology provides a predictable and controlled environment for automated parsing. This identification strategy follows the community best practices established by TEI Lex-0 (Burnard & Sperberg-McQueen, 2012; Tasovac et al., 2018), transforming the "hidden" hierarchies of the printed page into a fully navigable digital network.

## 4.4 Multiword Expressions (MWEs)

The identification and encoding of Multiword Expressions (MWEs) represent a significant challenge in the retro-digitisation of legacy dictionaries like *Al-Mawrid*. The linguistic nature of MWEs is traditionally defined by their statistical recurrence (Smadja, 1993) and semantic opacity, where the meaning of the unit cannot be directly derived from its individual components (Choueka, 1988; Sag et al., 2002). In print-based lexicography, MWEs are frequently "hidden" as subentries and lack explicit labeling, such as "idiom" or "collocation". This study addresses these issues by using a dual-standard approach to formalise these units as queryable lexical objects. Following TEI Lex-0 constraints, semantic information must reside within sense elements; thus, microstructural MWEs are uniformly represented as nested <relatedEntry> nodes to ensure structural validity and scalability. The encoding of a macrostructural MWE using explicit morphosyntactic declaration is shown below in Figure 5.

| **Full/Partial Entry** | cricket صَرَّارُ اللَّيْل: جُدْجُد |
|---|---|
| **LMF/TEI Lex-0 Encoding** | <entry xml:id="me-030303" type="mainEntry" xml:lang="ar"> <br> <form type="lemma"><orth>صَرَّارُ اللَّيْل</orth></form> <br> <gramGrp><gram type="mwe" value="N-N" /></gramGrp> <br> <sense n="se-01"><pc>:</pc><def>جُدْجُد</def></sense></entry> |

Figure 5: A Multiword Expression (MWE) headword.

For further variations, including transparent collocations and non-transparent idiomatic gemstone names, refer to Appendix D: Multiword Expressions.

### 4.4.1 Typology and Identification in Al-Mawrid

To resolve the structural inconsistencies of the source material, we categorise MWEs based on their placement (macrostructural vs. microstructural) and semantic transparency.

• Macrostructural MWEs: These function as primary headwords.

• Microstructural MWEs: These are embedded within other entries and are further distinguished by their transparency. Transparent MWEs (e.g., *ʿayn al-fiʿl*) are compositionally clear and lack formal definitions in the text. Conversely, Non-Transparent MWEs (e.g., *ʿayn al-shams* meaning "opal") are flagged by idiomatic definitions introduced by a colon (:) or by explicit cross-references (*rājiʿ*).

#### 4.4.2 Encoding Strategy: Motivations for Standard Alignment

In the digital encoding process, our methodology prioritises consistency and ISO-LMF interoperability over a mere reproduction of the print layout.

Macrostructural MWEs are encoded as <entry type="mainEntry">. To make the LMF layer explicit, we include a <gramGrp> containing a <gram type="mwe"> element with a @value attribute (e.g., "N-N" for Noun-Noun compounds). This provides the high-level syntactic specifications required by ISO 24613.

Microstructural MWEs are uniformly represented as nested <entry type="relatedEntry"> elements. The methodological motivation for this choice is governed by two factors:

1. TEI Lex-0 Constraints: Community guidelines require that semantic information (translations or definitions) reside within <sense> elements rather than <form> blocks. Because many *Al-Mawrid* MWEs contain their own translations, nesting them as entries ensures structural validity.

2. Scalability: Treating all MWEs as nested entries ensures a predictable, schema-compliant representation that supports automated extraction for NLP pipelines.

### 4.5 Cross-Referencing: Resolving Implicit Links

Cross-referencing in *Al-Mawrid* serves as a critical mechanism for reducing lexicographic redundancy, particularly when related words share a common root or form part of complex idiomatic compounds. In the original print format, these links are signaled by implicit cues, such as the Arabic keyword *rājiʿ* (راجع) or phrases like "refer to it in its place". The goal of our methodology is to transform these static, human-oriented indicators into machine-actionable URI targets using the TEI Lex-0 <ref> and <xr> elements. This formalisation is essential for differentiating general links from precise semantic equivalence in NLP applications. Figure 6 illustrates an example of resolving implicit pointers to multiple external senses.

| **Full/Partial Entry** | أَخَذَ (أَلْقَى، وَقَعَ) على عاتِقِه – راجعها في أماكنها ... |
|---|---|
| **LMF/TEI Lex-0** | <xr type="synonym"><pc>-</pc><lbl>راجعها في أماكنها</lbl><br><ref type="sense" target="#me-001625.se-22">أَخَذَ</ref><br><pc>،</pc><br><ref type="sense" target="#me-006840.se-19">أَلْقَى</ref><br><pc>،</pc><br><ref type="sense" target="#me-057995.se-17">وَقَعَ</ref></xr> |

Figure 6: Cross-Reference phrase resolving multiple targets.

Additional examples illustrating complete versus partial scoping and cross-references to idiomatic phrases are located in Appendix E: Cross-Referencing.

#### 4.5.1 Methodological Motivations for Encoding Strategies

To ensure structural rigour and interoperability, our encoding choices are governed by the specific constraints of the ISO-LMF and TEI Lex-0 frameworks:

- Structural Grouping: We utilize the <xr> (cross-reference) element as a functional wrapper for the entire referencing block. This choice is motivated by the need to group the typographic markers (such as the dash) with the functional labels and targets, maintaining the interpretive context of the original dictionary while ensuring the block is computationally distinct.

• Semantic Alignment: The relationship between the referring and referred entries is explicitly categorized using the @type="synonym" attribute within the <xr> element. This formalisation is essential for NLP applications, as it differentiates between a general "see also" link and a precise semantic equivalence.

• Target Disambiguation: A major challenge in *Al-Mawrid* is the existence of references that lack an explicit target word, such as "refer to them in their places". In these cases, our methodology requires an editorial intervention where the encoder manually identifies and assigns the appropriate @target URI. This process is vital for converting the dictionary into a functional network suitable for Linked Data frameworks.

4.5.2 Complete vs. Partial Scoping

A key distinction in our methodology is the treatment of the reference's scope. When a synonymy spans all senses of a lemma, the reference is encoded as complete, pointing to the top-level entry identifier. Conversely, when the synonymy is restricted to a specific meaning, the reference is partial (scoped), utilizing a target ID that points specifically to the relevant <sense> element. This layered approach ensures that the digital resource reflects the precise semantic alignment intended by the lexicographer. For a representative example of a cross-reference resolving multiple senses, see Figure 6. Further illustrations of MWE-related and complete references are available in the Appendix E.

4.6 Parenthesis Information: Resolving Structural Ambiguity

The encoding of parenthetical information in *Al-Mawrid* represents a critical methodological challenge, as these markers often collapse distinct types of information—ranging from grammatical metadata to unstructured semantic "clarity" cues—into a single typographic form. To address this, our methodology adopts the editorial view, treating parentheses as functional delimiters that must be disambiguated to ensure machine-readability and interoperability. Adopting the editorial view, we use the <gloss> element to preserve unstructured context that the author suggests is optional. This ensures descriptive annotations remain available for future automated extraction while maintaining a clean core lexical structure. The preservation of unstructured parenthetical context is demonstrated below in Figure 7.

| **Full/Partial Entry** | (في تَصْنيفِ الأَحْياء) عُوَيْلِم |
|---|---|
| **LMF/TEI Lex-0 Encoding** | <entry xml:id="me-001001" type="mainEntry" xml:lang="ar"><br><form type="lemma"><orth>عُوَيْلِم<br><gloss>(تَصْنيفِ في الأحْياء)</gloss></orth></form></entry> |

Figure 7: Unstructured parenthetical information.

A systematic presentation of structured versus unstructured parenthetical data types are provided in Appendix F: Parenthesis Information.

4.6.1 Typographic Classification and Logic

We distinguish between two types of parentheses used in the source material: square brackets [...] and braces (...). This distinction is central to our encoding strategy:

• Square Brackets [...]: These are functionally consistent in *Al-Mawrid*, exclusively containing domain or semantic field labels. These are mapped directly to the TEI element <usg type="domain">, transforming a visual cue into a queryable metadata tag.

• Braces (...): These are structurally ambiguous, containing a mix of structured (syntactic and semantic knowledge) and unstructured (contextual or illustrative) information.

### 4.6.2 Three-Tiered Encoding Strategy

To transform these braces into a standardised LMF/TEI Lex-0 format, we employ a three-tiered strategy governed by the degree of formal regularity in the text:

1. Transformation of Structured Metadata: When braces contain explicit linguistic cues (such as gender or part-of-speech indicators), the information is extracted and formalised using specific TEI Lex-0 elements like <gramGrp> or <lbl>. This decision is motivated by the ISO-LMF objective of making implicit linguistic knowledge explicit for NLP pipelines.

2. Transformation of Semi-Structured Information: Some parenthetical info—such as brief definitions or synonyms—is transformed into related entries or subsenses, ensuring that semantic relationships are not "hidden" within the text.

3. Preservation of Unstructured Content: Remaining unstructured text, which the original author noted as optional for "clarity," is preserved within the <gloss> element. This preserves the original lexicographic context for human readers while signalling to machine-parsers that the content is a descriptive annotation rather than a formal attribute.

### 4.6.3 Narrative Example: Disambiguating Context

The necessity of this editorial intervention is evidenced in entries like *ʿamīd* (Figure 27), where parenthetical info such as *(ḥizb, jamāʿa, etc.)* provides the semantic scope for the definition "chief." By encoding these as <gloss> elements within the <def> block, we maintain the hierarchical integrity of the sense while allowing future NLP tools to extract specific semantic domains.

## 4.7 Synonymy: Modelling Semantic Equivalence

The treatment of synonymy in the digital *Al-Mawrid* is grounded in the semantic theory of Cruse (1986), which defines synonyms as lexical items that share a central semantic core and can be substituted in context without altering a sentence's truth value. In a legacy bilingual resource, synonymy is not merely a linguistic relationship but a structural challenge; synonyms appear in multiple forms, ranging from explicit authorial markers to implied definitions embedded within explanatory text.

Guided by Cruse (1986), we treat synonymy as a structural challenge requiring transformation into a standardised semantic layer. To satisfy LMF requirements for high-level semantic specifications, each distinct synonymous defining term is nested within a <gloss> element inside the definition. This allows machine-parsers to identify each gloss as a discrete lexical unit for cross-linguistic mapping. Figure 8 provides a representative example of synonymous terms isolated within the definition block.

| **Full/Partial Entry** | **عَجِيبَة:** أَعْجُوبَة، مُعْجِزَة |
|---|---|
| **LMF/TEI Lex-0 Encoding** | <sense n="se-1"><pc>: </pc><br><def><gloss>أُعْجُوبَة</gloss><pc>،</pc><br><gloss>مُعْجِزَة</gloss></def></sense> |

Figure 8: (Partial) Definitional Synonyms.

Methodological proofs for header-level synonyms and explicitly marked equivalence are detailed in Appendix G: Synonymy.

### 4.7.1 Methodological Motivations for Encoding

Our methodology seeks to transform these varied print manifestations into a standardised semantic layer that supports both human readability and automated NLP tasks. The choice of specific TEI Lex-0 elements is motivated by the need to distinguish between pure definitions and lexical equivalents:

• Definitional Synonyms: When synonyms function as defining terms within an explanation, they are encoded within the <def> element. However, to satisfy the LMF requirement for high-level semantic specifications, each distinct synonymous term is nested within a <gloss> element. This structural choice is motivated by the need to facilitate sense-alignment; by isolating the synonym from the surrounding prose, we enable machine-parsers to identify the gloss as a discrete lexical unit suitable for cross-linguistic mapping.

• Header-Level Synonyms: In instances where synonymous lemmas appear together in a subentry header, our methodology employs <form type="variant" subtype="synonym">. This allows us to distinguish between simple orthographic variations and true semantic alternatives, preserving the lexical view of the entry while providing the metadata necessary for refined dictionary searches.

• Explicit Cross-Referencing: For synonyms marked by the authorial keyword *rājiʿ* (راجع), we utilize the <xr type="synonym"> attribute. The motivation here is to transition the dictionary from a static print resource into a functional sense-axis model in line with ISO 24613 (LMF). This enables the digital resource to support complex querying where a single search can retrieve a network of semantically related terms.

By formalising these implicit and explicit cues, the methodology ensures that the digital *Al-Mawrid* maintains its original lexicographic depth while becoming a harmonised, queryable resource that resolves the inconsistencies of the print medium.

## 4.8 Syntactic and Semantic Knowledge Modelling

This section explores the morpho-syntactic depth of Al-Mawrid, resolving scattered cues into an explicit model. Motivation: Explicit morphological markers (e.g., ج for plural) are mapped to standardised attributes in <gramGrp> to ensure FAIR compliance. This transition is essential for making the dictionary's internal logic computationally accessible without heuristic interpretation by the end-user. The Arabic components of the entries are exceptionally rich in implicit cues; a quantitative study by Fayed et al. (2014) highlights this lexicographic depth, offering a baseline for our mark-up of overt and inferred information

The following illustration demonstrates the methodological formalisation of a plural inflection into a structured computational model, a process necessitated by the need to transform scattered morphological cues into explicitly queryable data categories. By mapping these implicit linguistic markers to standardised XML tags, the methodology ensures that the dictionary's internal logic remains accessible to automated NLP pipelines without requiring heuristic interpretation by the end-user. This transition provides the morphological depth required to resolve the implicit relationships of the Arabic root-based system into a machine-actionable format.

| **Full/Partial Entry** | **عَجِيبَة** (ج عَجَائِب) |
|---|---|
| **LMF/TEI Lex-0 Encoding** | `<form type="lemma"><orth>عَجِيبَة</orth></form><form type="inflected"><pc>(</pc><lbl>ج</lbl><orth>عَجَائِب</orth><pc>)</pc><gramGrp><gram type="number" value="plural" /></gramGrp></form>` |

Appendix H: Syntactic and Semantic Knowledge Modelling contains comprehensive examples covering maṣdar markers and explicit POS tags.

#### 4.8.1 Disambiguating Explicit and Implicit Cues

Our methodology treats these cues as foundational data for Arabic NLP applications. The decision to formalise these cues is motivated by the need to resolve the structural inconsistencies typical of legacy print dictionaries, which often hide critical linguistic data within the phrasing of an entry.

• Explicit Cues: These are mapped directly to standardized attributes. For instance, grammatical labels (e.g., ج for plural) are encoded within the <gramGrp> element using the <gram> sub-element. This choice is driven by the ISO-LMF requirement for low-level specifications, ensuring that every overt linguistic marker becomes a queryable data category.

• Implicit Cues: The Arabic components of *Al-Mawrid* are exceptionally rich in inferred data, as highlighted by Fayed et al. (2014). Our encoding strategy involves an editorial intervention to transform these contextual patterns into explicit XML tags. This process is not merely a technical task but a methodological necessity to ensure that the dictionary's internal logic is accessible to computational tools without requiring heuristic interpretation by the end-user.

#### 4.8.2 Standardisation and Research Motivation

To support the study's dual-standard contribution, we align these features with both TEI Lex-0 and LMF. While TEI provides the vocabulary for the serialization (e.g., <usg> for domain labels), the LMF layer provides the conceptual framework for representing the underlying lexical objects. This alignment ensures that the resulting resource is not only structured but also interoperable with modern digital humanities infrastructures.

By systematising these syntactic and semantic markers, we move away from the "typographic view" of the printed page and toward a "lexical view" that prioritises semantic content. This transition is essential for making the dictionary a FAIR-compliant resource that serves as a robust model for other Arabic legacy dictionaries. Refer to Table 11 and Table 12 in Appendix G and Appendix H for the categorisation of these features, and Figures 40–55 for the practical application of markup for both overt and inferred information.

### 4.9 Homonym Modelling: Resolving Grammatical Ambiguity

A significant challenge in retro-digitising *Al-Mawrid* is the treatment of headwords that function across multiple grammatical categories without clear visual separation in the print edition. To resolve this, our methodology adopts a hierarchical nesting strategy for homonyms, transforming a single typographic entry into multiple, computationally distinct lexical objects.

Retro-digitising headwords that serve multiple grammatical roles requires a hierarchical nesting strategy. By marking these roles as type="homonymicEntry", we prioritise fine-grained linguistic analysis. This structure ensures each role is assigned a unique identity, allowing NLP pipelines to isolate and query specific grammatical behaviours independently. The treatment of headwords with dual noun/adjective functions is illustrated in Figure 9.

| **Full/Partial Entry** | (أَثِر : أَنَانِيّ (اسم / صفة |
|---|---|
| **LMF/TEI Lex-0** | `<entry xml:id="me-000777" ...><form type="lemma"><orth>أَثِر</orth></form> <entry xml:id="ho-01" type="homonymicEntry"><sense n="se-01"><def>أَنَانِيّ</def><gramGrp><gram type="POS" value="N"/></gramGrp></sense></entry> ... </entry>` |

Figure 9: Homonymic nesting for fine-grained analysis.

The full XML serialisation of complex homonymic entries is illustrated in Appendix I: Homonym Modelling

4.9.1 Methodological Motivation for Nested Entries

The decision to use nested <entry> elements, rather than a flat list of parts-of-speech (POS), is motivated by the need for microstructural precision. By marking these nested components with the @type="homonymicEntry" attribute, we ensure that each grammatical role—such as a lemma functioning as both a noun and an adjective—is assigned its own unique identity within the machine-readable file.

This approach is a strategic intervention that prioritises fine-grained linguistic analysis:

• Syntactic Clarity: Each nested entry includes an explicit <gramGrp> element, transforming the dictionary's often implicit POS cues into formal metadata.

• Computational Disambiguation: This structure allows NLP pipelines to isolate and query specific grammatical roles independently, ensuring the digital resource complies with the TEI Lex-0 baseline for interoperability.

4.9.2 Representative Application

The necessity of this structure is clearly evidenced in the entry for ʾaṯir (أَثِر). In the print edition, this lemma is listed once but serves two distinct functions. Our methodology treats these as two distinct entries (ho-01 for the noun and ho-02 for the adjective) nested within the primary lemma container. This ensures that the semantic definitions for each role remain logically separated and tied to their respective syntactic markers. (See Figure 46 in Appendix H for the complete XML representation of homonymic nesting for "أَثِر".)

4.10 Illustrative Examples and Contextual Cues

In legacy lexicography, illustrative examples and context words are essential for disambiguating the semantic boundaries of headwords, compounds, and idioms. Within the *Al-Mawrid*, these elements provide the syntactic and pragmatic framework necessary to understand typical usage in both the source and target languages. Our methodology prioritises the functional separation of these components to ensure that usage guides are computationally distinct from core definitions, thereby enhancing the resource's utility for Natural Language Processing (NLP) tasks.

Usage guides are essential for disambiguating semantic boundaries. Motivation: We employ the <cit type="example"> element to ensure that syntactic context markers (subjects/objects) are addressable via different XML paths than core definitions. This functional separation facilitates more precise automated analysis of the dictionary's semantic structure. Figure 10 encodes a verb usage phrase involving subject-object cues.

| **Full/Partial Entry** | هَمَّ (ـتِ الأُمُّ الطِّفْلَ) |
|---|---|
| **LMF/TEI Lex-0** | <entry xml:id="me-056491" type="mainEntry" xml:lang="ar"><form type="lemma"><orth>هَمَّ</orth></form><cit type="example"><quote><pc>(</pc> ـتِ الأُمُّ الطِّفْلَ<pc>)</pc>هَمَّ</quote></cit></entry> |

Figure 10: Functional separation of Illustrative Usage.

For further examples distinguishing usage sentences from context markers, see Appendix J: Illustrative Examples and Contextual Cues.

### 4.10.1 Modelling Illustrative Usage

Illustrative examples frequently appear as complete sentences or phrases designed to demonstrate a lemma's typical application. To preserve these usage guides without conflating them with the definition itself, we employ the TEI Lex-0 <cit type="example"> element. By encapsulating the example text within a <quote> block, we transform a visual cue into a structured citation that reflects the editorial view of the dictionary as a stream of tokens.

For instance, the entry for *hammama* (to lullaby) includes the phrase *(ـتِ الأُمُّ الطِّفْلَ)*, which provides the necessary syntactic subjects and objects for the verb. Encoding this as a structured citation ensures that the syntactic context is preserved in a machine-readable format while remaining distinct from the translation equivalent.

### 4.10.2 Contextual Annotation and Glossing

Distinct from illustrative sentences are context words, which often appear in parentheses to signal the specific circumstances under which a headword is used. These cues typically provide modifiers for nouns or subjects for verbs but do not constitute a full usage example.

The motivation for our encoding strategy is to distinguish these "clarity markers" from formal semantic attributes. We map these markers to the <gloss> element. A representative example is found in the entry *wāfaqa* (to coincide), where the parenthetical phrase *(في تارِيخٍ مُعَيَّن)* ("on a certain date") acts as a context marker rather than a full illustrative sentence.

By formalising this distinction, the methodology ensures that contextual cues and illustrative content are addressable via different XML paths, facilitating more precise automated analysis of the dictionary's internal semantic structure.

## 4.11 Translation Equivalence and Semantic Granularity

The encoding of English translation equivalents is a primary step in transforming *Al-Mawrid* from a static reference into a machine-actionable bilingual resource. Our methodology prioritises cross-linguistic alignment and the resolution of convergent meanings by mapping the dictionary's typographic markers (commas and semicolons) to hierarchical XML structures. The final step in creating a machine-actionable resource is the mapping of target-language data. Our methodology prioritises the segmentation of semicolon-delimited translation clusters into nested <sense> elements that function as distinct subsenses. Each resulting node is assigned a unique and stable identifier. This transformation of a flat list into a layered semantic model supports the fine-grained disambiguation required for modern NLP tasks. Figure 11 layers sense hierarchy for shades of convergent meaning.

| **Full/Partial Entry** | sin, wrongdoing; offense, fault, misdeed; iniquity; guilt إِثْم |
|---|---|
| **LMF/TEI Lex-0** | <sense n="se=01"><br><sense n="su-01"><br><cit type="translationEquivalent" xml:lang="en"><br><form><orth>sin</orth></form></cit><pc>,</pc><br><cit type="translationEquivalent" xml:lang="en"><br></cit><pc>,</pc><form><orth>wrongdoing</orth></form><br></cit></sense><pc>;</pc><br><sense n="su-02"><br><cit type="translationEquivalent" xml:lang="en"><br><form><orth>offense</orth></form></cit><pc>,</pc><br><cit type="translationEquivalent" xml:lang="en"><br></cit><pc>,</pc><form><orth>fault</orth></form><br></cit><cit type="translationEquivalent" xml:lang="en"> |

```xml
</cit><pc>,</pc><form><orth>misdeed</orth></form>
</cit></sense><pc>;</pc>
<sense n="su-03"><cit type="translationEquivalent" xml:lang="en">
<form><orth>iniquity</orth></form></cit><pc>,</pc></sense>
<pc>;</pc><sense n="su-04">
<cit type="translationEquivalent" xml:lang="en">
<form><orth>guilt</orth></form></cit><pc>,</pc></sense>
</sense>
</entry>
```

Figure 11: Hierarchical Translation Equivalence.

Additional tiered translation models and free-text equivalents are available in Appendix K: Translation Equivalence and Semantic Granularity.

4.11.1 Structural Modelling of Equivalents

To maintain interoperability within the TEI Lex-0 framework, we employ the <cit> (cited quotation) element as the formal container for all target-language data. This decision is motivated by the need to distinguish between different types of translation data:

• Formal Equivalents: Direct translation equivalents are marked with @type="translationEquivalent". Within these blocks, the <form> and <orth> elements provide a structured orthographic representation, ensuring the data is suitable for multilingual digital lexicons.

• Free-Text Translations: In instances where the translation is descriptive rather than a direct equivalent, the methodology substitutes <form> with the <quote> element (using @type="translation"). This intervention preserves the editorial view of the text while signalling to machine-parsers that the content is a semantic explanation rather than a lexical headword.

4.11.2 Resolving Ambiguity through Subsenses

A critical challenge in *Al-Mawrid* is the use of punctuation to signal semantic shades. In the print edition, comma-separated words indicate synonymous equivalents, while semicolons separate terms with "convergent meanings" or slightly different connotations.

To resolve this structural ambiguity, our methodology implements a layered sense hierarchy:

• Comma-delimited groups are treated as synonymous and stored within a single sense.
• Semicolon-delimited groups are parsed into distinct subsenses that are encoded by nested <sense> elements; each identified by a unique @n attribute.

This systematic use of the @n attribute—further detailed in our referencing framework—ensures that every shade of meaning is assigned a stable identifier. By formalising these visual boundaries, the methodology transforms a flat list of words into a structured semantic model that supports fine-grained linguistic analysis and NLP disambiguation tasks. (For a visual representation of this tiered translation encoding, see Figure 11).

## 5. Results and Discussion

This section evaluates the outcomes of the encoding process, focusing on the scientific weight and practical utility of the resulting resource. Rather than merely recapping the encoding steps, we provide a critical reflection on the limitations of TEI Lex-0 when applied to Arabic morphology and derivation, as well as the manual effort required for heuristic interpretation. We present evidence-based indicators of success—including structural coverage and navigational systematicity—and discuss the resource's potential for Linked Data export and integration into automated NLP pipelines.

This study successfully transformed the complex, implicit structures of *Al-Mawrid* into a machine-readable format that harmonises the ISO-LMF lexical view with TEI Lex-0 serialisation. Unlike a mere technical summary, this evaluation differentiates between the technical outcomes and the methodological challenges of applying Western-centric standards to the highly derivational nature of Arabic lexicography.

### 5.1 Quantitative Evaluation and Resource Coverage

To address the need for evidence-based indicators within the constraints of commercial licensing, we incorporate baseline statistics from Fayed et al. (2014) regarding the dictionary's overall coverage and linguistic variety. The methodology successfully captured the full hierarchy of the dictionary's macrostructure (~35,000 headwords) and microstructure (~60,000 subentries). Empirical analysis of a representative sample (the letter Ayn, comprising 4.6% of the volume) provides the following evidence-based indicators of the encoding's success:

- Structural Parsing Accuracy: Analysis of the microstructure's punctuation regularity reveals a structural parsing accuracy of 91%. The remaining 9% error rate was primarily due to "in-dictionary" inconsistencies, such as missing colons or mismatched headwords in the original print edition.
- Extraction Performance: Evaluation of the information extraction rules demonstrates high performance for semantic relations. Synonyms achieved a precision of 85% and a recall of 98%, validating the "Editorial View" approach to punctuation-driven parsing.
- Navigational Systematicity: The implementation of a prefix-based referencing scheme (e.g., me-, se-, su-) resolved the visual ambiguities of the print edition, transforming it into a predictable digital network where every lexical node is uniquely addressable.

### 5.2 Critical Reflection: Limitations and Framework Deviations

Applying TEI Lex-0 to a legacy Arabic resource revealed several representational "friction points" and necessitated specific methodological interventions:

- Implicit "Open Set" Relations: A significant portion of Arabic lexical knowledge is implicit; for example, 77% of hyponym/hypernym (HYP) relations in *Al-Mawrid* are "open set," embedded in natural language phrasing (genus-differentia) rather than explicit keywords as recorded by Fayed et al. (2014). TEI Lex-0 lacks internal mechanisms to automatically handle these, requiring expert "heuristic induction" to encode them as <xr type="hypernymy">.
- Morpho-Syntactic Gaps: *Al-Mawrid* lacks formal part-of-speech (POS) tags in over 99% of entries. To bridge this gap, we used LMF high-level specifications to guide the interpretation of scattered morphological cues (e.g., *maṣdar*) into explicit <gramGrp> blocks.
- MWE Metadata Constraints: Under TEI Lex-0, semantic data must reside within a <sense> element to maintain structural validity. Consequently, even compositionally transparent Multiword Expressions (MWEs) were encoded as nested <relatedEntry> nodes to ensure their translations were schema-compliant.

### 5.3 Manual Effort and Scalability

While rule-based segmentation handled the primary headword-sense divisions, human expertise was required for disambiguating non-transparent MWEs and unstructured parenthetical information. For instance, determining if a bracketed phrase was a domain label (<usg>) or a contextual gloss (<gloss>) required linguistic expertise that currently exceeds basic rule-based extraction. However, the current methodology creates a "Gold Standard" dataset that can be used to train future automated classifiers, significantly enhancing the scalability of the workflow for other Arabic legacy dictionaries.

### 5.4 Outlook: NLP Integration and Linked Data

The systematic assignment of unique @xml:id attributes facilitates the conversion of the dictionary into dereferenceable URIs for the Linguistic Linked Open Data (LLOD) cloud. By aligning the XML structures with the OntoLex-Lemon model, *Al-Mawrid* can be interlinked with other resources like Arabic WordNet or global classification systems like DBpedia. Furthermore, by resolving semicolon-delimited sense ambiguities into distinct nested <sense> nodes, the resource is now prepared for immediate integration into automated NLP pipelines, such as HPSG grammar systems or semantic disambiguation tools.

## 6. Conclusion and Outlook

The final section synthesises the paper's findings into three distinct areas of contribution: resource, methodological, and conceptual. We summarise how the digital *Al-Mawrid* serves as a novel resource for the Arabic-speaking world and how our dual-standard workflow offers a reusable model for other legacy dictionaries. Finally, we provide an outlook on future work, including the planned automation of extraction tools and the open-access publication of the resource to enhance the standardisation and interoperability of Arabic lexicography in the digital age.

This study has successfully transformed the *Al-Mawrid* Arabic-English dictionary from a static print resource into a standardised, machine-readable digital infrastructure. By implementing a dual-standard framework—aligning ISO-LMF with TEI Lex-0—we have demonstrated that the structural complexities of legacy Arabic lexicography can be resolved without compromising representational integrity.

### 6.1 Resource Contribution: The Digital Al-Mawrid

The primary output of this work is the first standardised digital encoding of *Al-Mawrid*, a widely used bilingual reference previously under-represented in the digital sphere.

- Pilot Scope and Coverage: The methodology was validated across the dictionary's complete macro- and microstructure, capturing diverse entries ranging from single-word lemmas to complex Multiword Expressions (MWEs).
- Navigational Indicators: Novelty is evidenced by our systematic referencing scheme, which assigned unique, persistent identifiers (e.g., me-000000 through me-999999) to every lexical node, transforming the dictionary into a predictable, queryable network suitable for NLP pipelines.

### 6.2 Methodological Contribution: A Reusable Workflow

We provide a reusable encoding workflow that addresses the specific challenges of retro-digitising Arabic legacy resources.

- Ambiguity Resolution: Our methodology successfully parsed implicit print structures, such as semicolon-delimited translation groups, into explicit XML <sense> and nested <sense> elements (se-01, su-01), thereby enhancing data granularity.
- Dual-Standard Mapping: A significant contribution is the explicit alignment between LMF's abstract lexical model and TEI Lex-0's serialisation, providing a blueprint for maintaining interoperability across different standardisation paradigms.
- Ultimately, this study provides a validated architectural blueprint for the digital preservation of the broader Arabic lexicographical tradition, ensuring that legacy print heritage is transformed into interoperable components of the global Linguistic Linked Open Data (LLOD) cloud.

6.3 Conceptual Contribution: Shifting Perspectives

The conceptual novelty of this research lies in its shift from a "typographic view" (preserving the print layout) to a "lexical view" (extracting underlying linguistic data).
• Standardisation of Arabic Complexity: By categorising MWEs into a typology based on semantic transparency (Transparent vs. Non-Transparent) and encoding them as distinct <relatedEntry> nodes, we have provided a linguistically grounded solution for representing idiomatic Arabic within Western-centric encoding standards.
• FAIR Principles: This study advances the FAIR (Findable, Accessible, Interoperable, and Reusable) publication of Arabic resources, bridging the gap between traditional lexicographic traditions and the requirements of the global Linguistic Linked Open Data (LLOD) community.

6.4 Future Directions

Building upon this foundation, future work will focus on the automation of heuristic decisions and the full-scale LOD export of the resource to link Al-Mawrid with broader Arabic ontological networks. Specifically, we intend to address the following areas to enhance the scalability and impact of the digital resource:

• Advanced Automation and Extraction: While the current methodology relies on manual heuristic interpretation to resolve structural inconsistencies, this phase serves as a necessary "Gold Standard" for training future tools. Future iterations will focus on training machine learning classifiers to automate the extraction and validation of parenthetical information and domain labels. These tools will specifically target the formalisation of "open set" implicit relations, such as the 77% of hyponymic links that currently exceed rule-based extraction capabilities.

• Semantic Web and LLOD Integration: To explore the potential of linked data technologies in depth, the next logical step is the serialisation of these XML structures into RDF using the OntoLex-Lemon model. The systematic prefix-based referencing scheme (e.g., *me-* for entries, *se-* for senses) provides the necessary scaffolding for dereferenceable URIs. This will enable the resource to participate in the global Linguistic Linked Open Data (LLOD) cloud, interlinking with resources such as Arabic WordNet and global classification systems like DBpedia or Wikidata.

• Infrastructure and FAIR Accessibility: In alignment with FAIR principles, we plan to release the encoded dictionary under an Open Access licence. This ensures the Arabic NLP community has access to a standardised, machine-actionable bilingual resource suitable for immediate integration into automated pipelines, such as HPSG grammar systems or semantic disambiguation tools.

• Expansion to Legacy Lexicography: The techniques developed here provide a reusable workflow that can be extended to other large-scale legacy Arabic dictionaries. By applying this "Editorial View" to the broader tradition of Arabic lexicography, we aim to transform static print heritage into a robust, interoperable digital infrastructure for modern linguistics

# APPENDICES

## Appendix A: 4.1 Entry Representation and Lemma Modelling

**Entry**: **Example 2**: The entry header is a lemma of a multiword expression (mwe)

| **Full/Partial Entry** | صَرَّارُ اللَّيْل: جُدْجُد cricket |
|---|---|
| **LMF/TEI Lex-0 Encoding** | <entry xml:id=”me-000003” type=”mainEntry” xml:lang=”ar”><br><form type=”lemma”><orth>صَرَّارُ اللَّيْل</orth></form><br><gramGrp resp=”#DiaaFayed #LaurentRomary”><br><gram type=”mwe” value=”N-N”/></gramGrp><br><sense n=”se-01”><pc>:</pc><def>جُدْجُد</def><br><cit type=”translationEquivalent” xml:lang=”en”><br><form><orth>cricket</orth></form></cit></sense></entry> |

Figure 12: a multiword expression header

**Entry**: **Example 3**: The header contains more than one lemma. The lemmas are orthographic variants.

| **Full/Partial Entry** | عَجُفَ، عَجِفَ: هَزَلَ |
|---|---|
| **LMF/TEI Lex-0 Encoding** | **<entry xml:id="me-000004" type="mainEntry" xml:lang="ar"><br><form type="lemma"><orth>عَجُف</orth><pc>،</pc><br><form type="variant"><orth>عَجِف</orth></form></form><br><sense n="se-01"><pc>:</pc><def>هَزَل</def></sense></entry>** |

Figure 13: a multi-lemma header

**Entry**: **Example 4**: The entry header has more than one lemma. The lemmas are orthographic variants.

| Full/Partial Entry | صُرْصُور، صُرْصُر: بِنْتُ وَرْدان cockroach<br>صُرْصُور، صُرْصُر: جُدْجُدْ، صَرَّارُ اللَّيْل cricket |
|---|---|
| LMF/TEI Lex-0 Encoding | <entry xml:id="me-000004" type="mainEntry" xml:lang="ar"><br><form type="lemma"><orth>صُرْصُور</orth><pc>،</pc><br><form type="variant" ><orth>صُرْصُر</orth></form></form><br><sense n="se-01"><pc>:</pc><def>بِنْتُ وَرْدان</def></sense><br><cit type="translationEquivalent" xml:lang="en"><br><form><orth> cockroach</orth></form></cit></sense></entry> |

Figure 14: a multi-lemma header

**Entry**: **Example 5**: The sub-entry header (صَرْخَة ، آخِرُ صَرْخَة) is composed of the lemma and a similar meaning phrase. In addition, when there are more than one defining words or phrases, we encode each defining phrase in an <gloss> element.

| Full/Partial Entry | صَرْخَة :صَيْحَة<br>صَرْخَة ، آخِرُ صَرْخَة: بِدْعَةٍ، صَرْعَة، مُوضَة |
|---|---|
| LMF/TEI Lex-0 Encoding | <entry xml:id="me-000006" type="mainEntry" xml:lang="ar"><br><form type="lemma"><orth>صَرْخَة</orth></form><br><sense n="se-01"><pc>:</pc><def>صَيْحَة</def></sense><br><sense n="se-02"><br><form type="lemma"><orth>صَرْخَة</orth></form><br><form type="lemma"><orth>آخِرُ صَرْخَة</orth></form> <pc>:</pc><def><br><gloss>مُوضَة</gloss><pc>،</pc<br><gloss>صَرْعَة</gloss><pc>،</pc><br><gloss>بِدْعَة</gloss></def></sense></entry> |

Figure 15: a multi-lemma or a multi-phrase header

**Entry**: **Example 6**: The sub-entry header (صَرَّحَ عَن or صَرَّحَ بِ) is composed of the lemma and a preposition.
The headword (صَرَّحَ) is collocated with prepositions (بِ) and (عَن).
The parenthesis information is encoded as an <gloss> element.

| Full/Partial Entry | أَعْلَنَ، جاهَرَ بِ :صَرَّح<br>صَرَّحَ: أَوْضَحَ، أَظْهَر<br>صَرَّحَ بِ: أَجازَ، رَخَّص<br>(صَرَّحَ عَن (دَخْلِهِ أو مُمْتَلَكاتِهِ الخاضِعَةِ لِلضَّرِيبَةِ أو الرَّسْم |
|---|---|
| LMF/TEI Lex-0 Encoding | <sense n="se=03"><form type="lemma"><orth>صَرَّح</orth></form><br><gramGrp><gram type="collocate">ب</gram></gramGrp></sense><br><sense n="se=04"><br><form type="lemma"><orth>صَرَّح</orth></form><gramGrp><br><gram type="collocate">عن</gram></gramGrp><br><gloss>( دَخْلِهِ أو مُمْتَلَكاتِهِ الخاضِعَةِ لِلضَّرِيبَةِ أو الرَّسْم )</gloss></sense> |

Figure 16: a prepositional phrase header

## Appendix B: 4.2 Labels, Punctuation, and Semantic Marking

Table 7: Summary Table for labels and abbreviations

| Arabic labels | | |
|---|---|---|
| **Label** | **Translation** | **Encoding** |
| [رياضة]، [فلك]، [قضاء]، [اقتصاد]، [سينما]، [جغرافيا] ، [سياسَة] ، [أَحياء] ، [طب] ، [فلسفة] ، [دين]، [كِيمياء]، ... | Sports, Astronomy, Judiciary, Economics, Geography, Politics, Biology, Medicine, Physiology, Religion, Chemistry | <pc>[</pc><br><usg type=”domain”><br>رياضة</usg><pc>]</pc> |
| (نبات)زُعْرُور<br>(حيوان)ثَيْتَل | plant<br>animal | <lbl>نبات</lbl><br><lbl>حيوان</lbl> |
| /إلخ/ | /Etc./ | <lbl>إلخ</lbl> |
| /راجع/<br>/راجعها في مكانها/<br>/راجعها في أماكنها/ | /refer/<br>/refer it in its place/<br>/refer them in their places/ | <lbl>راجع</lbl><br><lbl>راجعها في مكانها</lbl><br><lbl>راجعها في أماكنها</lbl> |
| /ج/ | /Plural/ | <lbl>ج</lbl> |
| /مفردها/ | /Singular/ | <lbl>مفردها</lbl> |
| /ضد/, /لا/, /غير/ | /Antonym/ | <lbl>ضد</lbl>,<lbl>لا</lbl>, <lbl>غير</lbl> |
| /إسم/ | /Noun / | <lbl>إسم</lbl> |
| /صفة/ | /Adjective/ | <lbl>صفة</lbl> |
| /أو/ | /or/ | <lbl>أو</lbl> |
| /و/ | /and/ | <lbl>و</lbl> |
| /مصدر / | /infinitive/ | <lbl>مصدر</lbl> |
| /طائِفَةٌ مِن/ | /Class of / | <lbl>طائِفَةٌ مِن</lbl> |
| /الفَصِيلَة/ | /family/ | <lbl>الفَصِيلَة</lbl> |

Table 8: Summary Table for labels and abbreviations

| Arabic Abbreviations | | |
|---|---|---|
| **Abbreviation** | **Translation** | **Encoding** |
| /ج=جَمْع/ | /Pl. = plural/ | <lbl>ج</lbl> |
| /إلخ=إلى آخره / | /Etc. = Etcetera/ | <lbl>إلخ</lbl> |
| /نفس = علم نفس/ | /Psych. = Psychology/ | <pc>[</pc><br><usg type=”domain”><br>نفس</usg><pc>]</pc> |
| /أحياء = علم أحياء / | /Biology = Biology/ | <pc>[</pc><br><usg type=”domain”> أحياء<br></usg><pc>]</pc> |
| /نبات = علم النبات/ | /plant = Botany/ | <lbl>نبات</lbl> |

Table 9: Summary Table for punctuations, labels and abbreviations.

| Arabic punctuations and labels | | |
|---|---|---|
| colon /:/ | what comes after them is a definition, interpretation, or explanation of the material before them. | <pc>:</pc> |
| comma /،/ | separates synonyms or two phrases that have the same meaning or word has multiple spelling variations. | <pc>،</pc> |
| square bracket /[ ]/ | contains one or more semantic domain labels | <pc>[</pc><br><lbl>لغة</lbl><br><pc>]</pc> |
| parentheses /( )/ | ▪ Brackets ( ) are used in many cases, most of which are intended to increase clarity. Accordingly, it is possible, in general, to omit what is placed in parentheses, that is, not to read it.<br>▪ Used in many cases, most of which are intended to increase clarity.<br>▪ The brackets are also used to denote the type or gender of a noun.<br>▪ Sometimes used in the cross-reference to determine the referenced main or branch entry word(words) | <pc>)</pc><br><pc>(</pc> |
| Ellipse /…/ | | <pc>…</pc> |
| Question mark /?/ | | <pc>؟</pc> |
| Exclamation mark /!/ | | <pc>!</pc> |
| dash /-/ | Before the cross-reference labels | <pc>-</pc> |

Table 10: Summary Table for punctuations, labels and abbreviations.

| English punctuations and labels | | |
|---|---|---|
| colon /,/ | separates synonymous English words | <pc>,</pc> |
| semicolon /;/ | separates English words with slightly different connotations or between shades of convergent meanings, but without being synonymous. | <pc>;</pc> |
| parentheses /( )/ | parentheses can be deleted or retained, because the parentheses indicate the permissibility of using the word in more than one form or include another word that is placed besides it or is synonymous with an adjacent word. for example: | <pc>)</pc><br><pc>(</pc> |
| Etc. /etc./ | | <lbl>Etc.</lbl> |
| Ellipse /…/ | | <pc>…</pc> |
| Question mark /?/ | | <pc>?</pc> |
| Exclamation mark /!/ | | <pc>!</pc> |
| Plural / Pl./ | Pl.=plural | |

## Appendix C: 4.3 Lexical Structure and Reference Modelling

```
<entry type="mainEntry" xml:id=”me-000001”>
   <entry type="homonymicEntry" xml:id=”he-000001”>
        <sense>
          <sense n=”se-01”>
             <sense n=”su-01”></sense>
             <sense n=”su-02>
                <cite xml:lang="en"></cite>
                <cite xml:lang="en"></cite>
             </sense>
          </sense>
          <sense n=”se-02”>
             <sense n=”su-01”></sense>
          </sense>
        </sense>
        <entry type="relatedEntry" xml:id=”re-000001”>
                          …
        </entry>
        <entry type="relatedEntry" xml:id=”re-000002”>
                          …
        </entry>
   </entry>
   <entry type="homonymicEntry" xml:id=”he-000002”>
   </entry>
</entry>
```

```
<entry type="relatedEntry" xml:id=”re-000001”>
   <sense>
        <sense n=”se-01”>
           <sense n=”su-01”></sense>
                 …
        </sense>
        <sense n=”se-02”>
           <sense n=”su-01”></sense>
           <sense n=”su-02”></sense>
        </sense>
   </sense>
</entry>
<entry type="relatedEntry" xml:id=”re-000002”>
   <sense>
      <sense n=”se-01”>
         <sense n=”su-01”></sense>
         <sense n=”su-02”>
           <cite xml: lang="en"></cite>
           <cite xml: lang="en"></cite>
         </sense>
         <sense n=”su-03”></sense>
      </sense>
    </sense>
</entry>
```

Figure 17: Figure 8: Macrostructure, Microstructure and Referencing system.

## Appendix D: 4.4 Multiword Expressions

**MWEs: Example 1:** The lemma of the entry is a MWE of two words.

| Full/Partial Entry | cricke صَرَّارُ اللَّيْل: جُدْجُد |
|---|---|
| LMF/TEI Lex-0 Encoding | <entry xml:id="me-030303" type="mainEntry" xml:lang="ar"><br><form type="lemma"><orth>صَرَّارُ اللَّيْل</orth></form><br><gramGrp resp="#DiaaFayed #LaurentRomary"><br><gram type="mwe" value="N-N"/></gramGrp><br><sense n="se-01"><pc>:</pc><def>جُدْجُد</def></sense></entry> |

Figure 18: a MWE lemma

**MWEs: Example 2:** The lemma of the entry is a MWE of two words. The header contains a semantic field or domain label.

| Full/Partial Entry | apraxia [طب] عَمَهٌ حَرَكِيّ |
|---|---|
| LMF/TEI Lex-0 Encoding | <entry xml:id="me-036984" type="mainEntry" xml:lang="ar"><br><form type="lemma"><orth>عَمَهٌ حَرَكِيّ</orth></form><br><gramGrp resp="#DiaaFayed #LaurentRomary"> <gram type="mwe" value="NN-JJ"></gram></gramGrp><br><sense n="se-01"> <usg type="domain" value="Medicine"><br><pc>[</pc>طب<pc>]</pc></usg><br><cit type="translationEquivalent" xml:lang="en"> <form> <orth>apraxia</orth><br></form> </cit></sense></entry> |

Figure 19: a MWE lemma with Semantic Field Label.

**MWEs: Example 3:** The two subentries are MWE s. The meaning of the collocation can be directly derived from its individual words. These are called transparent polylexical units.

| Full/Partial Entry | أَحَاطَ بِ: أَحْدَقَ بِ، طَوَّقَ<br>أَحَاطَ بِدائِرَةٍ أو بِشَكْلٍ هَنْدَسِيّ<br>أَحَاطَ بِ: شَمَلَ<br>أَحَاطَ بِهِ، أَحَاطَ بِهِ عِلْمًا: عَلِمَ بِ، اِطَّلَعَ على<br>أَحَاطَ بِهِ، أَحَاطَ بِهِ عِلْمًا: فَهِمَ<br>أَحَاطَهُ عِلْمًا بِ: أَعْلَمَهُ بِ، أَطْلَعَهُ على<br>أَحَاطَهُ بِالرِّعَايَةِ أو العِنَايَةِ أو الاهْتِمام – راجع رَعَى، عُنِيّ بِ، اِهْتَمَّ بِ |
|---|---|
| LMF/TEI Lex-0 Encoding | <sense n="se-02"><form type="collocations"><br><form type="collocation"><orth><br><ref type="form" scope="currentEntry" value="أَحَاطَ بِ"/><br><seg>دائِرَةٍ<lbl>أو</lbl>شَكْلٍ هَنْدَسِي</seg><br></orth><gramGrp><gram type="mwe" value="V-P-N-A"/><br></gramGrp></form></form></sense><br>....<br><sense n="se-06"><form type="collocations"><br><form type="collocation"><orth><br><ref type="form" scope="currentEntry" value="أَحَاطَهُ بِ"/><br><seg>اِلرِّعَايَة<lbl>أو</lbl>اِلعِنَايَة<lbl>أو</lbl>الاهْتِمام</seg><br></orth><gramGrp><gram type="mwe" value="V-N-P-N"/><br></gramGrp></form></form></sense> |

Figure 20: a transparent MWE.

**MWEs: Example 4:** The two subentries are MWE s. The meaning of the collocation can be directly derived from its individual words. These are called transparent polylexical units.

| Full/Partial Entry | عَيْنِيّ: ذُو عَلَاقَةٍ بِالعَيْن ...<br>حَقٌّ عَيْنِيّ، حُقُوقٌ عَيْنِيَّة right(s) in rem<br>مُسَاهَمَةٌ عَيْنِيَّة contributions in kind |
|---|---|
| LMF/TEI Lex-0 Encoding | <sense n="se=06"><br><form type="collocations"><br><form type="collocation"> <orth><seg>حَق</seg><ref type="oRef"/></orth><br><gramGrp><gram type="mwe" value="N-A"/></gramGrp><br><cit type="translation" xml:lang="en"> <quote xml:lang="en">right in rem</quote></cit> </form><pc>،</pc><br><form type="collocation"> <orth><seg>حُقُوق</seg><ref type="oRef"/><seg>ة</seg></orth><br><gramGrp><gram type="mwe" value="N-A"/></gramGrp></form><br><cit type="translation" xml:lang="en"> <quote xml:lang="en">rights in rem</quote></cit></form></form></sense><br><sense n="se=07"><br><form type="collocations"><form type="collocation"><orth><br><seg>مُسَاهَمَة</seg><ref type="oRef"/><seg>ة</seg></orth></form><br><gramGrp><gram type="mwe" value="N-A"/></gramGrp></form><br><cit type="translation" xml:lang="en"> <quote xml:lang="en">contributions in kind</quote></cit></sense> |

Figure 21: a transparent MWE.

**MWEs: Example 5:** transparent polylexical unit.

| Full/Partial Entry | عَيْنُ الفِعْل [لغة] second radical of a verb |
|---|---|
| LMF/TEI Lex-0 Encoding | <sense n="se-18"><form type="collocations"><form type="collocation"><br><orth><ref type="oRef"/><seg>الفِعْل</seg></orth><br><gramGrp resp="#DF #LR"><gram type="mwe" value="N-N"/></gramGrp></form></form> |

| | <usg type="domain" value="Language"><pc>[</pc>لغة<pc>]</pc></usg> <cit type="translation"> <quote xml:lang="en">second radical of a verb</quote></cit></sense> |
|---|---|

Figure 22: LMF/TEI-Lex0 code of a transparent MWE with semantic field label.

**MWEs: Example 6:** Encoding of a transparent MWE and semantic field.

| Full/Partial Entry | اسْمُ عَيْنٍ [لغة] concrete noun |
|---|---|
| LMF/TEI Lex-0 Encoding | <sense n="se-20"><form type="collocations"><form type="collocation"> <orth><seg>اسْم</seg><ref type="oRef"/></orth> <gramGrp resp="#DF #LR"><gram type="mwe" value="N-N"/></gramGrp></form></form> <usg type="domain" value="Language"><pc>[</pc>لغة<pc>]</pc></usg> <cit type="translationEquivalent" xml:lang="en"> <form><orth>concrete noun</orth></form></cit></sense> |

Figure 23: LMF/TEI-Lex0 code of a transparent MWE with semantic field label.

**MWEs: Example 7:** Encoding a non-transparent sense-related MWE.

| Full/Partial Entry | عَيْنُ الشَّمْسِ: أُوبَال (حَجَرٌ كَرِيم) opal |
|---|---|
| LMF/TEI Lex-0 Encoding | <sense n="se-17"> <entry xml:id="re-377377" type="relatedEntry" xml:lang="ar"> <form type="lemma"><orth> عَيْنُ الشَّمْسِ</orth></form> <gramGrp resp="#DF #LR"><gram type="mwe" value="N-N"/></gramGrp> <sense n="se-01"><pc>:</pc><def>أُوبَال<gloss>(حَجَرٌ كَرِيم) </gloss></def> <cit type="translationEquivalent" xml:lang="en"><form><orth>opal</orth></form></cit></sense></entry></sense> |

Figure 24: LMF/TEI-Lex0 code of non-transparent sense-related MWEs.

**MWEs: Example 8:** Encoding a non-transparent sense-related MWE.

| Full/Partial Entry | عَيْنُ الهِرّ: حَجَرٌ نِصْفُ كَرِيم cat's-eye |
|---|---|
| LMF/TEI Lex-0 Encoding | <sense n="se-19"> <entry xml:id="re-333433" type="relatedEntry" xml:lang="ar"> <form type="lemma"><orth> عَيْنُ الهِر</orth></form> <gramGrp resp="#DF #LR"><gram type="mwe" value="N-N"/></gramGrp> <sense n="se-01"><pc>:</pc><def>حَجَرٌ نِصْفُ كَرِيم</def> <cit type="translationEquivalent" xml:lang="en"><form><orth>cat's-eye</orth> </form></cit> </sense></entry> </sense> |

Figure 25: LMF/TEI-Lex0 code of non-transparent sense-related MWEs.

**MWEs: Example 9:** Encoding lexicographically non-transparent entry-related MWE.

| Full/Partial Entry | an eye for an eye, tit for tat العَيْنُ بالعَيْن |
|---|---|
| LMF/TEI Lex-0 Encoding | <entry xml:id="me-334140" type="relatedEntry" xml:lang="ar"> </orth></form> العَيْنُ بالعَيْن<form type="lemma"><orth> <gramGrp resp="#DF #LR"><gram type="mwe" value="N-P-N"/></gramGrp> <sense n="se-01"> <cit type="translation" xml:lang="en"><quote>an eye for an eye</quote> |
| | </cit><pc>,</pc><br><cit type="translation" xml:lang="en"><quote>tit for tat</quote></cit></sense></entry> |

Figure 26: LMF/TEI-Lex0 code of non-transparent entry-related MWEs.

**MWEs: Example 10:** Encoding lexicographically non-transparent entry-related MWE.

| Full/Partial Entry | بِالعَيْنِ المُجَرَّدَة with the naked (or unaided) eye |
|---|---|
| LMF/TEI Lex-0 Encoding | <entry xml:id="me=334141" type="relatedEntry" xml:lang="ar"><br><form type="lemma"><orth>بِالعَيْنِ المُجَرَّدَة</orth></form><br><gramGrp resp="#DF #LR"><gram type="mwe" value="P-N-N"/></gramGrp><br><sense n="se-01"><cit type="translation" xml:lang="en"><br><quote>with the naked <pc>(</pc>or unaided<pc>)</pc> eye<br></quote></cit> </sense></entry> |

Figure 27: LMF/TEI-Lex0 code of non-transparent entry-related MWEs.

# Appendix E: 4.5 Cross-Referencing: Resolving Implicit Links

**4.5.1 Example 1:** Cross-Reference Phrase without Targets.

- The dash symbol "-" is encoded using the <pc> (punctuation) element.
- The keyphrase used for the cross-reference, such as "راجعها في أماكنها", is encoded using the <lbl> (label) element.
- The keyphrase "راجعها في أماكنها" is not followed by any <ref> element or specific reference target.
- In this instance, the <ref> elements were added by the author of the paper for clarification.
- The absence of explicit reference targets after the phrase "راجعها في أماكنها" is considered a deficiency in the original dictionary structure.
- Reference targets, when present, are composed of two parts: the entry identifier and the sense identifier.

| Full/Partial Entry | عاتِق: كَتِف<br>عاتِقُ المِيزانِ أو المِحْراثِ ...<br>أَخَذَ (أَلْقَى، وَقَعَ) على عاتِقه – راجعها في أماكنها ...<br>أَخَذَ : تَنَاوَلَ، نَالَ ...<br>أَخَذَ على عاتِقِه أو نَفْسِه (معنى 22) ...<br>أَلْقى :رَمَى، قَذَفَ ...<br>أَلْقَى على عاتِقِهِ كَذَا (معنى 19) ...<br>وَقَعَ :سَقَطَ ...<br>وَقَعَ على عاتِقِه (معنى 17) ...<br>أَخَذَ: تَنَاوَلَ، نَالَ ...<br>أَخَذَ على عاتِقِه أو نَفْسِه (معنى 22) |
|---|---|
| LMF/TEI Lex-0 Encoding | <xr type="synonym"> <pc>-</pc> <lbl>راجعها في أماكنها</lbl><br><ref type="sense" target="#me-001625.se-22">أَخَذ</ref><pc>،</pc><br><ref type="sense" target="#me-006840.se-19">أَلْقَى</ref><pc>،</pc><br><ref type="sense" target="#me-057995.se-17">وَقَعَ</ref></xr> |

Figure 28: Cross-Reference phrase without targets.

**4.5.2 Example 2:** Cross-Reference to Multiple Senses

- The dash “-” is encoded using the <pc> element, and the keyword "راجع" is marked using the <lbl> element.
- The keyword "راجع" refers to three different senses from separate entries.
- Both the dash and the keyword are included within an <xr> element, representing the cross-reference structure.
- The attribute @type="synonym" within the <xr> element indicates that the relationship between the subentry and the referenced senses or entries is one of synonymy.

| **Full/Partial Entry** | ... طَوَّق ،بِأَحْدَق :أَحَاطَ بِ<br>... أَحَاطَهُ بِالرِّعَايَةِ أو العِنَايَةِ أو الاهْتِمام . راجع رَعَى، عُنِيَ بِ، اِهْتَمَّ بِ<br>رَعَى: عُنِيَ بِ، حَفِظَ<br>رَعَى: تَعَهَّدَ، نَمَّى<br>رَعَى :حافَظَ على، اِحْتَرَمَ، تَقَيَّدَ بِ. راجع راعَى<br>رَعَى: حَكَمَ، سادَ، طُبِّقَ على<br>رَعَى الماشِيَةَ أو القَطِيعَ<br>رَعَتِ الماشِيَةُ<br>**رَعَى** (النُّجُومَ إلخ): راقَبَ – راجع راعَى ...<br>عُنِيَ بِ :اِهْتَمَّ بِ...<br>**اِهْتَمَّ بِ**: قَلِقَ<br>اِهْتَمَّ بِ: اِغْتَمَّ<br>اِهْتَمَّ بِ: عُنِيَ بِ<br>اِهْتَمَّ بِ: بالَى بِ، حَفَلَ بِ |
|---|---|
| **LMF/TEI Lex-0 Encoding** | <xr type="synonym"> <pc>.</pc> <lbl>راجع</lbl><br><ref type="entry" target="#me-003700">رَعَى</ref> <pc>،</pc><br><ref type="entry" target="#me-004155">عُنِيَ بِ</ref><pc>،</pc><br><ref type="entry" target="#me-002784">اِهْتَمَّ بِ</ref></xr> |

Figure 29: Cross-reference phrase for more than one refers.

**4.5.3 Example 3:** Cross-Reference to a Multiword Expression (MWE)

- In this example, the keyword "راجع" refers to the phrase "على أَثَرِ كَذَا", which is treated as a multiword expression (MWE) and appears as a subentry under the headword "أَثَر".
- Occasionally, the author assists users by including the headword of the referred entry in parentheses—e.g., "(أَثَر)"—immediately after the cross-reference.
- This parenthetical inclusion is rare and represents an exception to the typical implicit referencing pattern used in the dictionary.

| **Full/Partial Entry** | أَثَر: تَأْثِير، مَفْعُول<br>على أَثَرِ كَذَا، في أَثَرِ كَذَا<br>على أَثَرِ كَذَا (أَثَر في إِثْرِ كَذا . راجع ،إِثْر) |
|---|---|
| **LMF/TEI Lex-0 Encoding** | <xr type="synonym"> <pc>.</pc> <lbl>راجع</lbl><br><ref type="sense" target="#me-000760.se-02">على أَثَرِ كَذَا</ref><br><pc>(</pc>أَثَر<pc>)</pc></xr> |

Figure 30: Cross-Reference to a Multiword Expression (MWE).

**4.5.4 Example 4:** Complete Cross-Reference.

- In Figure 9, the cross-reference is complete, meaning there is full semantic alignment between the referring and referred entries.
- Specifically, the headword "فَهَّمَ" and the referenced word "أَفْهَمَ" share all their senses without discrepancy.
- This allows the reference to be encoded with a clear @target attribute pointing to the full set of senses under the referenced entry.

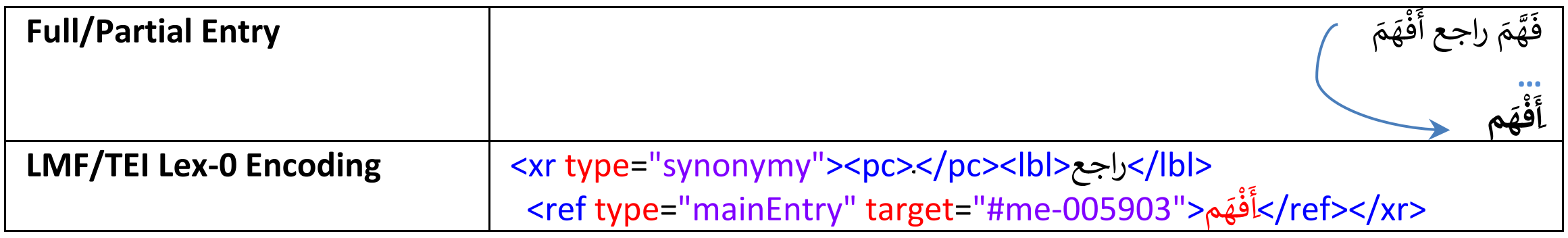

| Full/Partial Entry | فَهَّمَ راجع أَفْهَمَ<br>...<br>أَفْهَمَ |
|---|---|
| LMF/TEI Lex-0 Encoding | <xr type="synonymy"><pc>.</pc><lbl>راجع</lbl><br><ref type="mainEntry" target="#me-005903">أَفْهَمَ</ref></xr> |

Figure 31: Complete Cross-Reference (refer to all the meanings).

#### 4.5.5 Example 5: Partial Cross-Reference

- In Figure 10, the cross-reference points to a specific sense rather than the full lexical entry.
- The referring subentry "فَرَضَ: حَز" links to a particular sense of the entry "فَرَّض" using a target such as #me-38951.
- This example illustrates scoped synonymy, where only one meaning is shared across otherwise distinct entries.

| Full/Partial Entry | فَرَضَ: حَزَّ. راجع فَرَّضَ<br>...<br>فَرَّضَ: حَزَّ |
|---|---|
| LMF/TEI Lex-0 Encoding | **<xr type="synonymy"> <pc>.</pc> <lbl>راجع</lbl><br><ref type="mainEntry" target="#me-038951.se-01">فَرَّض</ref></xr>** |

Figure 32: FPartial Cross-reference (only one meaning).

## Appendix F: 4.6 Parenthesis Information: Resolving Structural Ambiguity

**Parenthesis Information: Example: 1:**

| Full/Partial Entry | **عُوَيْلِم** (في تَصْنيفِ الأَحْياء) |
|---|---|
| LMF/TEI Lex-0 Encoding | **<entryxml:id="me-001001" type="mainEntry" xml:lang="ar"><br><form type="lemma"><orth>عُوَيْلِم<gloss>( في تَصْنيفِ<br>الأحْياء)</gloss></orth></form></entry>** |

Figure 33: Code of unstructured parenthetical information.

**Parenthesis Information: Example: 2**

| Full/Partial Entry | **عَلامَة** (مَدْرَسِّية، عَلى إمْتِحانٍ إلخ) |
|---|---|
| LMF/TEI Lex-0 Encoding | <entryxml:id="**me-001002**" type="**mainEntry**" xml:lang="**ar**"><br><form type="**lemma**"> <orth>عَلامَة<gloss>(مَدْرَسِّية، عَلى إمْتِحانٍ إلخ)</gloss><br></orth></form></entry> |

Figure 34: Code of unstructured parenthetical information.

**Parenthesis Information: Example: 3**

| Full/Partial Entry | **عَلامَةُ تَرْقِيم** (كَالنُّقْطَةِ و الفاصِلَة) |
|---|---|
| LMF/TEI Lex-0 Encoding | **<entryxml:id="me-001003" type="mainEntry" xml:lang="ar"><br><form type="lemma"><orth>عَلامَةُ تَرْقِيم<gloss>(كَالنُّقْطَةِ و الفاصِلَة)</gloss><br></orth></form></entry>** |

Figure 35: Code of unstructured parenthetical information.

**Parenthesis Information: Example: 4**

| Full/Partial Entry | عَمِيد: سَيِّد، رَئِيس (حِزْب، جَمَاعَةٍ إلخ) |
|---|---|
| LMF/TEI Lex-0 Encoding | **<entry xml:id="me-001004" type="mainEntry" xml:lang="ar"><br><form type="lemma"><orth>عَمِيد</orth></form><br><sense n="se-01"><pc>:</pc><def><gloss>سَيِّد</gloss><pc>،</pc><br><gloss>رَئِيس<gloss>(حِزْب، جَمَاعَةٍ إلخ)</gloss></gloss></def></sense></entry>** |

Figure 36: Code of unstructured parenthetical information.

**Parenthesis Information: Example: 5**

| Full/Partial Entry | غَيَّمَ (بِتِ السَّماءُ) – راجع غامَ (بِتِ السَّماءُ) |
|---|---|
| LMF/TEI Lex-0 Encoding | <entry xml:id="me-001005" type="mainEntry" xml:lang="ar"><br><form type="lemma"><orth>غَيَّم<gloss>(بِتِ السَّماء)</gloss></orth></form><br><sense n="se-01"><pc>-</pc><lbl>راجع</lbl><xr type="synonym"><br><ref type="sense" target="#me-343434">غام<gloss>( بِتِ<br>السَّماء)</gloss></ref></xr></sense></entry> |

Figure 37: Code of unstructured parenthetical information.

**Parenthesis Information: Example: 6**

| Full/Partial Entry | فَتَّشَ (شَخْصًا بِجَسِّ ثِيَابِه بَحْثًا عن سِلَاح) |
|---|---|
| LMF/TEI Lex-0 Encoding | <entry xml:id="me-001007" type="mainEntry" xml:lang="ar"><br><form type="lemma"> <orth>فَتَّش<gloss><br>(شَخْصًا بِجَسِّ ثِيَابِه بَحْثًا عن سِلَاح)</gloss></orth></form></entry> |

Figure 38: Code of unstructured parenthetical information.

**Parenthesis Information: Example: 7**

| Full/Partial Entry | فِعْلِيًّا - راجع فِعْلًا، بالفِعْل (فِعْل) |
|---|---|
| LMF/TEI Lex-0 Encoding | <entry xml:id="me-001010" type="mainEntry" xml:lang="ar"><br><form type="lemma"> <orth>فِعْلِيًّا</orth></form><br><sense n="se-01"><pc>-</pc><lbl>راجع</lbl><xr type="synonym"><br><ref type="entry" target="#me-001400">فِعْلًا</ref> <pc>،</pc><br><ref type="entry" target="#me-003255">بالفِعْل <gloss><br>(<gloss>(فِعْل)</gloss>)</gloss></ref> </xr></sense></entry> |

Figure 39: Code of unstructured parenthetical information.

# Appendix G: 4.7 Synonymy: Modelling Semantic Equivalence

# Appendix H: 4.8 Syntactic and Semantic Knowledge Modelling

**Syntactic and Semantic Knowledge: Example: 1**

| Full/Partial Entry | عَجِيبَة (ج عَجَائِب): أعْجُوبَة، مُعْجِزَة |
|---|---|
| LMF/TEI Lex-0 Encoding | <entry xml:id="me-035404" type="mainEntry" xml:lang="ar"><br><form type="lemma"><orth>عَجِيبَة</orth></form><form type="inflected"><br><pc>(</pc><lbl>ج<lbl/><orth>عَجَائِب</orth><pc>)</pc><br><gramGrp><gram type="number" value="plural"/></gramGrp><br></form><sense n="se-1"><pc>:</pc> <def><gloss>أعْجُوبَة</gloss><pc>،</pc><br><gloss>مُعْجِزَة</gloss></def></sense></entry> |

Figure 40: Syntactic and Semantic Knowledge.

**Syntactic and Semantic Knowledge: Example: 2**

| Full/Partial Entry | عَجَائِب (مفردها عَجِيبَة) - راجع عَجِيبَة |
|---|---|
| LMF/TEI Lex-0 Encoding | <entry xml:id="me-035317" type="mainEntry" xml:lang="ar"><br><form type="lemma"><orth>عَجَائِب</orth></form><form type="inflected"><br><pc>(</pc><lbl>مفردها</lbl><orth>عَجِيبَة</orth><pc>)</pc><br><gramGrp><gram type="number" value="singular"/></gramGrp><br></form><pc>-</pc><lbl>راجع</lbl><br><sense n="se-01"><xr type="inflected"><br><ref target="#me-035404" type="entry">عَجِيبَة</ref></xr><sense></entry> |

Figure 41: Syntactic and Semantic Knowledge.

**Syntactic and Semantic Knowledge: Example: 3**

| Full/Partial Entry | ّمُبْتَل :(نَدٍ (النَّدِي |
|---|---|
| LMF/TEI Lex-0 Encoding | <entry xml:id="me-054156" type="mainEntry" xml:lang="ar"><br><form type="lemma"><orth>نَدٍ</orth><br><gramGrp resp="#DF #LR"><gram type="pos"<br>value="N"/></gramGrp></form><br><form type="inflected"><pc>(</pc><orth>النَّدِي</orth><pc>)</pc><br><gramGrp resp="#DF #LR"><gram type="pos"<br>value="N"/></gramGrp></form><br><sense n="se-01"><pc>:</pc><def>مُبْتَل</def></sense></entry> |

Figure 42: Syntactic and Semantic Knowledge.

**Syntactic and Semantic Knowledge: Example: 4**

| Full/Partial Entry | وَسَط (أوْساط): دائِرَة (دَوَائِر)، مَحْفِل (مَحَافِل) |
|---|---|
| LMF/TEI Lex-0 Encoding | <entry xml:id="me-057506" type="mainEntry" xml:lang="ar"><form type="lemma"><br><orth>وَسَط</orth></form><form<br>type="inflected"><pc>(</pc><orth>أوْساط</orth><pc>)</pc><br><gramGrp resp="#DF #LR"><gram type="number" value="plural"/></gramGrp></form><br><sense n="se-3"><pc>:</pc><def>دائِرَة</def><form<br>ype="inflected"><pc>(</pc><orth>دَوَائِر</orth><pc>)</pc><br><gramGrp resp="#DF #LR"><gram type="number"<br>value="plural"/></gramGrp></form><pc>،</pc><br><def>مَحْفِل</def><form type="inflected"><pc>(</pc><orth>مَحَافِل</orth><pc>)</pc><br><gramGrp resp="#DF #LR"><gram type="number"<br>value="plural"/></gramGrp></form></sense></entry> |

Figure 43: Syntactic and Semantic Knowledge.

**Syntactic and Semantic Knowledge: Example: 5**

| Full/Partial Entry | راهَقَ: صارَ مُرَاهِقاً |
|---|---|
| LMF/TEI Lex-0 Encoding | <entry xml:id="me-26200" type="mainEntry" xml:lang="ar"><br><form type="lemma"><orth>راهَق</orth><br><gramGrp resp="#DF #LR"><gram type="pos"<br>value="V"/></gramGrp></form><br><form type="derivative"><lbl>صار</lbl><orth>مُرَاهِقاً</orth><br><gramGrp resp="#DF #LR"><gram type="pos"<br>value="N"/></gramGrp></form><br><sense n="se-01"><pc>:</pc><def>صارَ مُرَاهِقاً</def></sense></entry> |

Figure 44: Syntactic and Semantic Knowledge.

**Syntactic and Semantic Knowledge: Example: 6**

| Full/Partial Entry | خِلْطِيّ: مَنْسُوبٌ إلى الخِلْط |
|---|---|
| LMF/TEI Lex-0 Encoding | <entry xml:id="me-23559" type="mainEntry" xml:lang="ar"><br><form type="lemma"><orth>خِلْطِي</orth><br><gramGrp resp="#DF #LR"><gram type="pos"<br>value="ADJ"/></gramGrp></form><br><form type="derivative"><lbl> مَنْسُوب<br>إلى</lbl><orth>الخِلْط</orth><gramGrp resp="#DF #LR"><br><gram type="pos" value="N"/></gramGrp></form><br><sense n="se-01"><pc>:</pc><def>مَنْسُوبٌ إلى الخِلْط</def></sense></entry> |

Figure 45: Syntactic and Semantic Knowledge.

**Syntactic and Semantic Knowledge: Example: 7**

| Full/Partial Entry | أَثِر: أَنَانِيّ (صفة)<br>أَثِر: أَنَانِيّ (اسم ) |
|---|---|
| LMF/TEI Lex-0 Encoding | <entry xml:id="me-000777" type="mainEntry" xml:lang="ar"><br><form type="lemma"><orth>أَثِر</orth></form> |
| | <entry xml:id="ho-01" xml:lang="ar" type="homonymicEntry"><br><sense n="se-01"><pc>:</pc><br><def>أَنَانِي</def><pc>(</pc><lbl>اسم</lbl><pc>)</pc><br><gramGrp resp="#DF #LR"><gram type="POS" value="N"/><br></gramGrp></sense></entry><br><entry xml:id="ho-02" xml:lang="ar" type="homonymicEntry"><br><sense n="se-01"><pc>:</pc><br><def>أَنَانِي</def><pc>(</pc><lbl>صفة</lbl><pc>)</pc><br><gramGrp resp="#DF #LR"><gram type="POS" value="ADJ"/><br></gramGrp></sense></entry></entry> |

Figure 46: Syntactic and Semantic Knowledge.

**Syntactic and Semantic Knowledge: Example: 8**

| Full/Partial Entry | ساوَقَ: ناغَمَ، جَعَلَهُ مُتَسَاوِقاً |
|---|---|
| LMF/TEI Lex-0 Encoding | <entry xml:id="me-028719" type="mainEntry" xml:lang="ar"><br><form type="lemma"><orth>ساوَقَ</orth><gramGrp resp="#DF #LR"><br><gram type="pos" value="V"/></gramGrp></form><form type="derived"><br><lbl>جَعَلَه</lbl><orth>مُتَسَاوِقا</orth><gramGrp resp="#DF #LR"><gram type="pos"<br>value="ADJ"/><br></gramGrp></form><sense n="se-01"><pc>:</pc><def><gloss>ناغَم</gloss><br><pc>،</pc><gloss>جَعَلَهُ مُتَسَاوِقا</gloss></def></sense></entry> |

Figure 47: Syntactic and Semantic Knowledge.

**Syntactic and Semantic Knowledge: Example: 9**

| Full/Partial Entry | مُبَلِّط: فاعِل بَلَّط |
|---|---|
| LMF/TEI Lex-0 Encoding | <entry xml:id="me-045287" type="mainEntry" xml:lang="ar"><br><form type="lemma"><orth>مُبَلِّط</orth></form><br><gramGrp resp="#DF #LR"><gram type="pos" value="N"/></gramGrp><br><form type="derived"><gramGrp resp="#DF #LR"><gram type="pos"<br>value="V"/></gramGrp><br><orth>بَلَّط</orth></form><sense n="se-01"><pc>:</pc><def> فاعِل<br>بَلَّط</def></sense></entry> |

Figure 48: Syntactic and Semantic Knowledge.

**Syntactic and Semantic Knowledge: Example: 10**

| Full/Partial Entry | اِبْتِغاء: مَصْدَر اِبْتَغَى |
|---|---|
| LMF/TEI Lex-0 Encoding | <entry xml:id="me-000197" type="mainEntry" xml:lang="ar"><br><form type="lemma"><orth>اِبْتِغاء</orth></form><br><gramGrp resp="#DF #LR"><gram type="pos" value="N"/></gramGrp><br><form type="derived"><gramGrp resp="#DF #LR"><gram type="pos"<br>value="V"/></gramGrp><br><orth>اِبْتَغَى</orth></form><sense n="se-01"><pc>:</pc><def> مَصْدَر<br>اِبْتَغَى</def></sense></entry> |

Figure 49: Syntactic and Semantic Knowledge.

**Syntactic and Semantic Knowledge: Example: 11**

| Full/Partial Entry | تَصالَحَ: ضِدّ تَخاصَمَ |
|---|---|
| LMF/TEI Lex-0 Encoding | <entry xml:id="me-25565" type="mainEntry" xml:lang="ar"><br><form type="lemma"><orth>تَصالَح</orth></form><br><gramGrp resp="#DF #LR"><gram type="POS" value="V"/></gramGrp><br><sense n="se-01"><pc>:</pc><def> ضِدّ تَخاصَم</def><br><xr type="antonymy" resp="#DF #LR"><br><ref target="#me-013006.se-01" ype="sense">تَخَاصَم</ref><br></xr></sense></entry> |

Figure 50: Syntactic and Semantic Knowledge.

**Syntactic and Semantic Knowledge: Example: 12**

| Full/Partial Entry | حُمَّى: حَرَارَة [طب] |
|---|---|
| LMF/TEI Lex-0 Encoding | <entry xml:id="me-20565" type="mainEntry"  xml:lang="ar"><br><form type="lemma"><orth>حُمَّى</orth></form><br><sense n="se-01"><pc>:</pc><def>حَرَارَة</def><pc>[</pc><br><usg type="domain"<br>value="medicine"><lbl>طب</lbl></usg><pc>]</pc></sense></entry> |

Figure 51: Syntactic and Semantic Knowledge.

**Syntactic and Semantic Knowledge: Example: 13**

| Full/Partial Entr | عابِد: مَنْ يَعْبُدُ |
|---|---|
| LMF/TEI Lex-0 Encoding | <entry xml:id="me-034848" type="mainEntry" xml:lang="ar"><br><form type="lemma"><orth>عابِد</orth><br><gramGrp resp="#DF #LR"><gram type="pos"<br>value="N"/></gramGrp></form><br><form type="derived"><orth>يَعْبُدُ</orth><br><gramGrp resp="#DF #LR"><gram type="pos"<br>value="V"/></gramGrp></form><br><sense n="se-01"><pc>:</pc><def>مَنْ يَعْبُدُ</def></sense></entry> |

Figure 52: Syntactic and Semantic Knowledge.

**Syntactic and Semantic Knowledge: Example: 14**

| Full/Partial Entry | طائِفَةٌ مِنَ النَّبَاتات :عارِيَاتُ البُزُور |
|---|---|
| LMF/TEI Lex-0 Encoding | <entry xml:id="me-034962"  type="mainEntry" xml:lang="ar"><br><form type="lemma"><orth> عارِيَاتُ البُزُور</orth></form> <sense n="se-01"><pc>:</pc><br><def> طائِفَةٌ مِنَ النَّبَاتات</def><xr type="hypernymy"><lbl>طائِفَةٌ مِن </lbl><br><ref type="entry" target="#me-053763">النَّبَاتات</ref></xr></sense></entry> |

Figure 53: Syntactic and Semantic Knowledge.

**Syntactic and Semantic Knowledge: Example: 15**

| Full/Partial Entry | زُرْقُطَة (حشرة) |
|---|---|
| LMF/TEI Lex-0 Encoding | <entry type="mainEntry"><form type="lemma"><orth>زُرْقُطَة</orth><br></form><xr type="hypernymy"><ref type="entry"><br><gloss>(حشرة)</gloss></ref></xr></entry> |

Figure 54: Syntactic and Semantic Knowledge.

**Syntactic and Semantic Knowledge: Example: 17**

| Full/Partial Entry | زُغْرُور(نبات) |
|---|---|
| LMF/TEI Lex-0 Encoding | <entry type="mainEntry"><form<br>type="lemma"><orth>زُغْرُور</orth><br></form><xr type="hypernymy"><ref type="entry"><br><gloss>(نبات)</gloss></ref></xr></entry> |

Figure 55: Syntactic and Semantic Knowledge.

Table 11: Information Types, Examples, and Encoding.

| Information Category | Information Type | explicit / implicit | Example | LMF/LEXO |
|---|---|---|---|---|
| *orthographic* | variation | explicit | عَجُفَ ،عَجِفَ: هَزَلَ | <form type="lemma"> |
| *morphologic* | inflection: number (singular) | explicit | عَجَائِب (مفردها عَجِيبَة) . راجع عَجِيبَة | <form type="variant"><orth></orth> |
| | inflection: number (plural) | explicit | عَجِيبَة (ج عَجَائِب): أَعْجُوبَة، مُعْجِزَة | </form> <form type="variant"><orth></orth></form></form> |
| | inflection: number (plural) | implicit | وَسَط (أوْساط): دائِرَة (دَوَائِر)، مَحْفِل (مَحَافِل) | <form type="inflected"> |
| | inflection: *active participles* (اسم فاعل) | explicit | هادٍ (الهادِي): مُرْشِد | <form type="inflected"> |
| | derivation: deverbal nouns (مصدر *maSdar*) | explicit | اِبْتِغاء: مَصْدَر اِبْتَغَى | <form type="inflected"> |
| | | implicit | مُبَلِّط: فاعِل بَلَّطَ | <form type="inflected"> |
| | | implicit | خِلْطِيّ: مَنْسُوبٌ إلى الخِلْط | <form type="derivation"> |
| | | implicit | راهَقَ: صارَ مُرَاهِقاً | <form type="derivation"> |
| | | implicit | ساوَقَ: ناغَمَ، جَعَلَهُ مُتَسَاوِقاً | <form type="derivation"> |
| *syntactic* | part-of-speech: noun | explicit | (أَثِر: أَنَانِيّ (صفة<br>)اسم) عالِم:عالِم | <form type="derivation"> |
| | part-of-speech: adj | explicit | أَثِر: أَنَانِيّ (اسم)<br>عالِمٌ بِ :عالِم (صفة) | <form type="derivation"> |
| *Illustrative example* | example | explicit | هَمَّمَ (ـتِ الأُمُّ الطِّفْلَ) | <gram type="pos" value="NN"> |
| *context gloss* | annotation, comment, … | explicit | صادَفَ (في تارِيخٍ مُعَيَّن) | <gram type="pos" value="ADJ"> |

Table 12: Information Types, Examples, and Encoding.

| | | | | |
|---|---|---|---|---|
| *semantic* | Semantic field (domain) | explicit | حُمَّى: حَرَارَة ]طب [ | <usg type="domain" value="medicine">طب</usg> |
| | antonyms | explicit | عاقّ :عَقُوق، ضِدّ بارّ<br>تَصالَحَ: ضِدّ تَخاصَمَ | <xr type="antonymy"> |
| | antonyms | explicit | اِتِّفَاقِيّ: لا إرَادِيّ | <ref target="", type=""> |
| | antonyms | explicit | ثابِت: غَيْرُ مُتَحَرِّكٍ | |
| | synonyms | implicit | قاتَلَ ،شاجَرَ :عارَكَ<br>عارِيَة ،عارِيَّة: قَرْض | |
| | | explicit | (إِثْرَ، فِي إِثْرِ كَذا . راجع على أَثَرِ كَذَا) أَثَر | <xr type="synonym" <ref target="" type="sense"></ref></xr> |
| | | explicit | فَهَّمَ راجع أَفْهَمَ | |
| | | implicit | فِكْرَةٌ (عِبَارَةٌ ،صِيْغَة (مُبْتَذَلَة | |
| | | implicit | طَرِيقَةٌ )أو أَسْلُوب (ـه في العَيْش | |
| | hypernym/hyponym relation: animal-of | explicit | خَيْلانِيّ: حَيَوانٌ مِنَ الخَيْلانِيّات | |
| | hypernym/hyponym relation: order-of | explicit | لاذَنَبِيّات: رُتْبَةٌ مِنَ البَرْمائِيّات | <xr type="hypernym" <ref type="entry"></ref></xr> |
| | hypernym/hyponym relation: class-of | explicit | عارِيَاتُ البُزُور: طائِفَةٌ مِنَ النَّبَاتات | <usg type="domain" value="medicine">طب</usg> |
| | hypernym/hyponym relation: flock-of | | رَفّ: جَمَاعَةٌ مِنَ الطَّيْرِ | <xr type="antonymy"> |
| | Meronym/Holonym relation: member-of | | نائِب: (عُضْو في) مَجْلِس النُّوَّاب | <ref target="", type=""> |
| | Meronym/Holonym relation: member-of | | زَمِيل: عُضْوٌ في جَمْعِيةٍ عِلْمِيَّة أو أَدَبِيَّة | |
| | Meronym/Holonym relation: member-of | | رَجُلُ عِصابَة، عُضْوٌ في عِصَاب | |
| | Meronym/Holonym relation: part-of | | عُضْو (مِنْ أَعْضاءِ الجِسْم) | <xr type="synonym" <ref target="" type="sense"></ref></xr> |
| | Meronym/Holonym relation: member-of | | عُضْو (في جَمْعِيَّة إلخ) | |
| | Meronym/Holonym relation: part-of | | فِقْرَة: جُزْء مِنْ (بَرْنامَج أو حَفْلَةٍ إلخ) | |
| | Meronym/Holonym relation: part-of | | قِسْط: جُزْءٌ مُقَسَّطٌ مِنْ دَيْن | |
| | Meronym/Holonym relation: part-of | | قَلَمُ السِّمَة: جُزْءٌ مِنْ مِدَقَّةِ الزَّهْرَةِ يَحْمِلُ السِّمَةَ | |
| | hypernym/hyponym relation: genus-of plant | implicit | زَنْبَقُ الماء (نبات) | <xr type="hypernym" <ref type="entry"></ref></xr> |
| | hypernym/hyponym relation: genus-of animals | implicit | أَبُو ذَقْن (حيوان) | <usg type="domain" value="medicine">طب</usg> |
| | human/profession: thematic, agent, actor | implicit | عابِد: مَنْ يَعْبُدُ | <xr type="antonymy"> |

## Appendix I: 4.9 Homonym Modelling: Resolving Grammatical Ambiguity

## Appendix J: 4.10 Illustrative Examples and Contextual Cues

| Full/Partial Entry | وافَقَ :صادَفَ (فِي تارِيخٍ مُعَيَّن) |
|---|---|
| LMF/TEI Lex-0 Encoding | <entry xml:id="me-056870" type="mainEntry" xml:lang="ar"><br><form type="lemma"><orth>وافَق</orth></form><gramGrp resp="#DF #LR"><br><gram type="POS" value="V"/></gramGrp><sense n="se-01"><pc>:</pc><br><def>صادَف</def><gloss>( فِي تارِيخ مُعَيَّن)</gloss></sense></entry> |

Figure 56: Examples and Context Words.

| Full/Partial Entry | هَمَّمَ (بتِ الأُمُّ الطِّفْلَ) |
|---|---|
| LMF/TEI Lex-0 Encoding | <entry xml:id="me-056491" type="mainEntry" xml:lang="ar"><br><form type="lemma"><orth>هَمَّم</orth></form><br><cit type="example"><quote> <pc>(</pc> تِ الأُمُّ<br>الطِّفْل<pc>)</pc>هَمَّم</quote></cit></entry> |

Figure 57: Examples and Context Words.

## Appendix K: 4.11 Translation Equivalence and Semantic Granularity

| Full/Partial Entry | sin, wrongdoing; offense, fault, misdeed; iniquity; guilt إِثْم |
|---|---|
| LMF/TEI Lex-0 Encoding | <sense n="se=01"><br><sense n="su-01"><br><cit type="translationEquivalent" xml:lang="en"><br><form><orth>sin</orth></form></cit><pc>,</pc><br><cit type="translationEquivalent" xml:lang="en"><br></cit><pc>,</pc><form><orth>wrongdoing</orth></form><br></cit></sense><pc>;</pc><br><sense n="su-02"><br><cit type="translationEquivalent" xml:lang="en"><br><form><orth>offense</orth></form></cit><pc>,</pc><br><cit type="translationEquivalent" xml:lang="en"><br></cit><pc>,</pc><form><orth>fault</orth></form><br></cit><cit type="translationEquivalent" xml:lang="en"><br></cit><pc>,</pc><form><orth>misdeed</orth></form><br></cit></sense><pc>;</pc><br><sense n="su-03"><cit type="translationEquivalent" xml:lang="en"><br><form><orth>iniquity</orth></form></cit><pc>,</pc></sense><br><pc>;</pc><sense n="su-04"><br><cit type="translationEquivalent" xml:lang="en"><br><form><orth>guilt</orth></form></cit><pc>,</pc></sense><br></sense><br></entry> |

Figure 58: Translations